\documentclass{article}

\newif\ifarXiv
\arXivtrue

\PassOptionsToPackage{numbers, compress, sort}{natbib}

\usepackage[final, eandd]{neurips_2026}

\usepackage{microtype}
\usepackage{graphicx}
\usepackage{booktabs}
\usepackage{multirow}
\usepackage{paralist}
\usepackage{subcaption}
\usepackage{todonotes}
\usepackage[hang, flushmargin]{footmisc}
\usepackage{silence}

\usepackage{hyperref}

\usepackage{amsmath}
\usepackage{amssymb}
\usepackage{mathtools}
\usepackage{amsthm}
\usepackage[english]{babel}

\usepackage{paralist}

\usepackage[separate-uncertainty=true,multi-part-units=single]{siunitx}

\usepackage[dvipsnames]{xcolor}

\usepackage{enumitem}
\usepackage{dsfont}
\usepackage{tcolorbox}
\usepackage{colortbl}
\usepackage{apptools}
\usepackage{titletoc}

\usepackage{thmtools}
\usepackage{thm-restate}
\usepackage{tikz}
\usepackage{tikz-3dplot}
\usepackage{wrapfig}
\usepackage{MnSymbol}
\usetikzlibrary{arrows.meta}
\usetikzlibrary{shapes}
\usetikzlibrary{calc, shapes.geometric}
\usepackage{svg}

\definecolor{OIOrange}    {RGB}{230, 159,   0}
\definecolor{OISkyBlue}   {RGB}{ 86, 180, 233}
\definecolor{OIGreen}     {RGB}{  0, 158, 115}
\definecolor{OIYellow}    {RGB}{240, 228,  66}
\definecolor{OIBlue}      {RGB}{  0, 114, 178}
\definecolor{OIVermillion}{RGB}{213,  94,   0}
\definecolor{OIRose}      {RGB}{204, 121, 167}

\definecolor{GLLow}  {RGB}{186,228,179}
\definecolor{GLow}  {RGB}{116,196,118}
\definecolor{GMid}  {RGB}{49,163,84}
\definecolor{GHigh} {RGB}{0,109,44}

\hypersetup{
    colorlinks,
    citecolor = OIGreen,
    urlcolor  = OIGreen,
    linkcolor = OIGreen,
}

\tcbset{
  boxsep     = 1.0mm,
  colframe   = OIGreen,
  colback    = OIGreen!10,
  left       = 1.0mm,
  right      = 1.0mm,
}

\declaretheorem[name=Theorem]{theorem}

\declaretheorem[name=Corollary]{corollary}

\usepackage[capitalize, nameinlink]{cleveref}

\DeclareMathOperator{\lk}{lk}

\title{No Triangulation Without Representation:\\
Generalization in Topological Deep Learning
}

\renewcommand{\thefootnote}{\fnsymbol{footnote}}

\author{Johannes S. Schmidt\footnotemark[1]\\
  Dept.\ of Informatics \\ 
  University of Fribourg \\
  Fribourg, Switzerland \\
  {\scriptsize\url{johannes.schmidt@unifr.ch}}\\
  \And
  Martin Carrasco\footnotemark[1]\\
  Dept.\ of Informatics \\ 
  University of Fribourg \\
  Fribourg, Switzerland \\
  {\scriptsize\url{martin.carrascocastaneda@unifr.ch}}\\
  \And
  Ernst Röell \\
  Inst.\ of AI for Health \\
  Helmholtz Munich \\
  Munich, Germany \\
  {\scriptsize\url{ernst.roell@helmholtz-munich.de}}
  \And
  Guy Wolf \\ 
  Dept.\ of Math.\ \& Statistics \\
  University of Montréal \\ 
  Montréal, Canada \\ 
  {\scriptsize\url{wolfguy@mila.quebec}}\\
  \And
  Nello Blaser\footnotemark[2]\\
  Dept.\ of Informatics \\ 
  University of Bergen \\
  Bergen, Norway \\
  {\scriptsize\url{nello.blaser@uib.no}}\\
  \And
  Bastian Rieck\footnotemark[2]\\
  Dept.\ of Informatics \\ 
  University of Fribourg \\
  Fribourg, Switzerland \\
  {\scriptsize\url{bastian.grossenbacher@unifr.ch}}\\
}

\begin{document}

\maketitle

\footnotetext[1]{These authors contributed equally to this work.}
\footnotetext[2]{These authors jointly supervised this work.}

\setcounter{footnote}{0}
\renewcommand{\thefootnote}{\arabic{footnote}}

\begin{abstract}
    Despite an ever-increasing interest in topological deep learning models that target higher-order datasets, there is no consensus on \emph{how} to evaluate such models.
This is exacerbated by the fact that topological objects permit operations, such as structural refinements, that are not appropriate for graph data.
In this work, we extend \texttt{MANTRA}, a benchmark dataset containing manifold triangulations, to a larger class of manifolds with more diverse homeomorphism types.  
We show that, contrary to prior claims, graph neural networks can
    \emph{saturate} the benchmark, while higher-order message passing
    methods lag behind.
However, we find that this is contingent on the right representation and feature assignment, emphasizing their importance in baseline models.
Hence, we present a novel evaluation protocol based on representational diversity and triangulation refinement.
Even when including refinement-like changes in the training data, existing models fail to generalize beyond the combinatorial structure of the data.
This points towards a research gap in developing models that understand topological structure independent of scale.
Our work thus provides the necessary scaffolding to evaluate future models and enable the development of topology-aware inductive biases.
\end{abstract}

\section{Introduction}

\emph{Topological deep learning}~(TDL) models aim to provide paradigms for learning on higher-order datasets, i.e., datasets that go beyond the dyadic relations captured in graphs~\citep{Hajij23b, Papamarkou24a}.
A prominent strand of TDL research is \emph{higher-order message passing}~(HOMP).
Originally developed for cell
complexes~\citep{bodnar2021cwnetworks, bodnar21weisfeiler, hajij2020cell}, HOMP has since been formalized and extended to the setting of combinatorial complexes~\citep{Hajij23a}, providing a general framework for learning on a variety of data modalities~\citep{Hajij23b}.
The field has thus developed a plethora of different models for addressing tasks on higher-order domains like cell complexes, CW complexes, or simplicial complexes, many of which are based on HOMP or variations thereof~\citep{Ballester24a, barsbey2025higherorder, Carrasco25a, Ebli20a, Roddenberry21a}, signal processing~\citep{Sardellitti21a, yang2025hodgeaware}, or integral transforms~\citep{Maggs24a, Rieck25a, Roell24a}.
Understanding higher-order data to the point of generalization is not only
relevant for topology \emph{per se} but also for applied mathematics or physics,
where lattice quantum field theory, for example, can be studied via space-time
triangulations~\citep{van_der_Duin_Loll_Schiffer_Silva_2026}.
However, higher-order data differs from graph data in crucial aspects.
Most importantly, the \emph{same} underlying space typically admits many combinatorial representations; that is, any manifold can be described by several triangulations~(cf.\ \cref{sec:Background}).
Models that only rely on combinatorial data will generalize poorly in such cases, necessitating novel model evaluation strategies for topological data.
\texttt{TopoBench}~\citep{telyatnikov2025topobench}, for instance, already
provides a large-scale benchmark and evaluation framework for TDL, but its
datasets are predominantly graphs \emph{lifted} to higher-order representations,
a procedure that does not reflect intrinsic properties of higher-order spaces~\citep{Rieck26a}.
As a consequence, model evaluations are effectively based on graph data, thus undermining the credibility of TDL models. This has led to criticism concerning the necessity of higher-order representations~\citep{peixoto2026graphsmaximallyexpressivehigherorder} and concerns about the expressivity of HOMP~\citep{eitan2025topological} or TDL~\citep{Ballester25b}.
The \texttt{MANTRA} dataset~\citep{ballester2025mantra} partially addresses this gap, consisting of \emph{inherently} higher-order objects, namely \emph{combinatorial manifolds}.
However, despite providing suitable \emph{data}, we find that the original \texttt{MANTRA} dataset has several shortcomings in its evaluation methodology, which we address in this work.

Our work makes the following \textbf{contributions}:
\begin{itemize}[left = 0pt, itemsep = 0.125em]
    \item We \emph{extend} the \texttt{MANTRA} dataset with a fully-characterized set of triangulations in dimensions $2$ and $3$, establishing \emph{Pachner moves}~\citep{Pachner1991}, i.e., local modifications preserving the homeomorphism type, as principled data augmentation and evaluation tools.\footnote{The dataset is publicly available at \url{https://github.com/aidos-lab/mantra}.}
\item We systematically investigate the impact of different
    \emph{representations} and \emph{encodings}~(positional and structural) on
    predictive performance, demonstrating that standard message-passing
    algorithms, given an appropriate data representation, \emph{outperform}
    higher-order models on this benchmark.\footnote{Code for performing such evaluations is available at \url{https://github.com/aidos-lab/mantra-pp-benchmarks}.}
\item For all datasets, we introduce \emph{refinement schemes} on the triangulations as a generalization ``stress test,'' revealing the fact that \emph{all} considered models \emph{fail} to generalize across refinements, even after seeing similar modifications in the training set, thus raising doubts about whether they learn topological structure or merely exploit combinatorial artifacts.
\end{itemize}
Taken together, our findings reveal that current TDL models are not genuinely
topological in the sense that they appear to learn combinatorial \emph{artifacts} rather
than topological \emph{structure}. This calls for a fundamental reassessment of
model design and evaluation. \begin{figure}[tbp]
    \centering
    \subcaptionbox{Simplicial complex $\mathcal{T} \cong S^2$\label{sfig:simplicial-complex}
    }{\begin{tikzpicture}[font=\small, line join=round, line cap=round]
    \begin{scope}[scale=1.25]
    \coordinate (v2) at ( 1.00,  0.00);
    \coordinate (v3) at (-0.50,  0.00);
    \coordinate (v4) at (0.25, -0.866);
    \coordinate (v1a) at (-1.50,  0.00);
    \coordinate (v1b) at ( 1.00, -1.5);
    \coordinate (v1c) at ( 1.00,  1.25);

    \draw[<->, dotted, thick] ($(v1a)!0.5!(v3) + (0, 0.1)$) to[out=60, in=150] ($(v1c)!0.5!(v3) + (-0.1, 0)$);
    \draw[<->, dotted, thick] ($(v1a)!0.5!(v4) + (0, -0.1)$) to[out=260, in=240] ($(v1b)!0.5!(v4) + (-0.05, -0.1)$);
    \draw[<->, dotted, thick] ($(v1c)!0.5!(v2) + (0.1, 0.0)$) to[out=20, in=30] ($(v1b)!0.5!(v2) + (0.1, 0)$);

    \fill[OIOrange!50] (v2)  -- (v3) -- (v4)  -- cycle;
    \fill[OISkyBlue!50]   (v1a) -- (v3) -- (v4)  -- cycle;
    \fill[OIGreen!50]  (v1b) -- (v2) -- (v4)  -- cycle;
    \fill[OIYellow!50]    (v1c) -- (v2) -- (v3)  -- cycle;

    \draw[very thick, black!80] (v2) -- (v3) -- (v4) -- cycle;
    \draw[very thick, black!80] (v1a) -- (v3)  (v1a) -- (v4);
    \draw[very thick, black!80] (v1b) -- (v2)  (v1b) -- (v4);
    \draw[very thick, black!80] (v1c) -- (v2)  (v1c) -- (v3);

    \node[fill=white, inner sep=1pt, font=\scriptsize] at ($(v2)!0.5!(v3)$) {$e_4$};
    \node[fill=white, inner sep=1pt, font=\scriptsize] at ($(v3)!0.5!(v4)$) {$e_5$};
    \node[fill=white, inner sep=1pt, font=\scriptsize] at ($(v2)!0.5!(v4)$) {$e_6$};
    \node[fill=white, inner sep=1pt, font=\scriptsize] at ($(v1a)!0.5!(v3)$) {$e_2$};
    \node[fill=white, inner sep=1pt, font=\scriptsize] at ($(v1a)!0.5!(v4)$) {$e_3$};
    \node[fill=white, inner sep=1pt, font=\scriptsize] at ($(v1b)!0.5!(v2)$) {$e_1$};
    \node[fill=white, inner sep=1pt, font=\scriptsize] at ($(v1b)!0.5!(v4)$) {$e_3$};
    \node[fill=white, inner sep=1pt, font=\scriptsize] at ($(v1c)!0.5!(v2)$) {$e_1$};
    \node[fill=white, inner sep=1pt, font=\scriptsize] at ($(v1c)!0.5!(v3)$) {$e_2$};

    \foreach \p in {v2,v3,v4,v1a,v1b,v1c} \filldraw[black] (\p) circle (0.028);
    \node[font=\small, right=1pt]        at (v2)  {$v_2$};
    \node[font=\small, above=1pt]        at (v3)  {$v_3$};
    \node[font=\small, below=1pt]        at (v4)  {$v_4$};
    \node[font=\small, left=1pt]         at (v1a) {$v_1$};
    \node[font=\small, below right=0.5pt] at (v1b) {$v_1$};
    \node[font=\small, above right=0.5pt] at (v1c) {$v_1$};

    \node[font=\small\bfseries] at ($0.333*(v2)+0.333*(v3)+0.334*(v4)$) {$\sigma_1$};
    \node[font=\small\bfseries]   at ($0.333*(v1a)+0.333*(v3)+0.334*(v4)$) {$\sigma_3$};
    \node[font=\small\bfseries]  at ($0.333*(v1b)+0.333*(v2)+0.334*(v4)$) {$\sigma_4$};
    \node[font=\small\bfseries]    at ($0.333*(v1c)+0.333*(v2)+0.334*(v3)$) {$\sigma_2$};

    \end{scope}
\end{tikzpicture} }
\subcaptionbox{Hasse diagram $\mathcal{H}$\label{sfig:hasse-diagram} 
    }{\begin{tikzpicture}[font=\small, line join=round, line cap=round, scale = 0.75,
  s2/.style = {circle,
               draw, very thick, black!80, inner sep=2pt,
               shape border uses incircle, 
               shape border rotate=180
               }]

\node[s2, rounded corners, fill=OIYellow!50,    font=\scriptsize, minimum width=08mm] (s2) at (-3.0, 2.0) {$\sigma_2$};
    \node[s2, rounded corners, fill=OISkyBlue!50,   font=\scriptsize, minimum width=08mm] (s3) at (-1.0, 2.0) {$\sigma_3$};
    \node[s2, rounded corners, fill=OIGreen!50,  font=\scriptsize, minimum width=08mm] (s4) at ( 1.0, 2.0) {$\sigma_4$};
    \node[s2, rounded corners, fill=OIOrange!50, font=\scriptsize, minimum width=08mm] (s1) at ( 3.0, 2.0) {$\sigma_1$};

\foreach \i/\x in {1/-2.5, 2/-1.5, 3/-0.5, 4/0.5, 5/1.5, 6/2.5}
        \node[draw, very thick, black!80, circle, fill=gray!15, font=\scriptsize, inner sep=2pt] (e\i) at (\x, 0.0) {$e_\i$};

\foreach \i/\x in {1/-1.5, 2/-0.5, 3/0.5, 4/1.5}
        \node[draw, very thick, black!80, circle, fill=black!10, font=\scriptsize, inner sep=2pt] (v\i) at (\x, -2.0) {$v_\i$};

\draw[black!80] (s1) -- (e4);  \draw[black!80] (s1) -- (e5);  \draw[black!80] (s1) -- (e6);
    \draw[black!80] (s2) -- (e1);  \draw[black!80] (s2) -- (e2);  \draw[black!80] (s2) -- (e4);
    \draw[black!80] (s3) -- (e2);  \draw[black!80] (s3) -- (e3);  \draw[black!80] (s3) -- (e5);
    \draw[black!80] (s4) -- (e1);  \draw[black!80] (s4) -- (e3);  \draw[black!80] (s4) -- (e6);

\draw[black!80] (e1) -- (v1);  \draw[black!80] (e1) -- (v2);
    \draw[black!80] (e2) -- (v1);  \draw[black!80] (e2) -- (v3);
    \draw[black!80] (e3) -- (v1);  \draw[black!80] (e3) -- (v4);
    \draw[black!80] (e4) -- (v2);  \draw[black!80] (e4) -- (v3);
    \draw[black!80] (e5) -- (v3);  \draw[black!80] (e5) -- (v4);
    \draw[black!80] (e6) -- (v2);  \draw[black!80] (e6) -- (v4);
\end{tikzpicture} }
\subcaptionbox{Dual graph $\mathcal{D} \cong K_4$\label{sfig:dual-graph}
    }{\begin{tikzpicture}[font=\small, line join=round, line cap=round, scale = 0.90,
  s2/.style = {circle, 
               draw, very thick, black!80, inner sep=2pt,
               shape border uses incircle}]

    \node[s2, fill=OIOrange!50, minimum size=5mm] (d1) at (0,2)    {$\sigma_1$};
    \node[s2, fill=OIYellow!50,    minimum size=5mm] (d2) at (-1.7,0)  {$\sigma_2$};
    \node[s2, fill=OISkyBlue!50,   minimum size=5mm] (d3) at (0,-1.2)  {$\sigma_3$};
    \node[s2, fill=OIGreen!50,  minimum size=5mm] (d4) at (1.7,0)   {$\sigma_4$};

    \draw[very thick, black!80] (d1) -- node[font=\scriptsize, fill=white, inner sep=1pt] {$e_4$} (d2);
    \draw[very thick, black!80] (d1) -- node[font=\scriptsize, fill=white, inner sep=1pt] {$e_5$} (d3);
    \draw[very thick, black!80] (d1) -- node[font=\scriptsize, fill=white, inner sep=1pt] {$e_6$} (d4);
    \draw[very thick, black!80] (d2) -- node[font=\scriptsize, fill=white, inner sep=1pt] {$e_2$} (d3);
    \draw[very thick, dotted, black!80] (d2) -- node[font=\scriptsize, fill=white, inner sep=1pt] {$e_1$} (d4);
    \draw[very thick, black!80] (d3) -- node[font=\scriptsize, fill=white, inner sep=1pt] {$e_3$} (d4);
\end{tikzpicture} }
\caption{
    Three representations of $S^2$, the $2$-sphere.  The simplicial complex and Hasse diagram contain the same information, while the dual graph only represents maximal simplex connectivity.
    }
    \label{fig:triangulation-representation}
\end{figure}

\section{Background: Simplicial complexes and combinatorial manifolds}
\label{sec:Background}

In this paper, we consider \emph{combinatorial manifolds}, which afford parsimonious representations by means of \emph{simplicial complexes}.
A simplicial complex $\mathcal{T}$ is a finite collection of simplices~(finite nonempty sets), that is closed under taking \emph{faces}~(subsets).
Given a simplex $\sigma = \{v_0, v_1, \dotsc, v_k\}$, we say that $\dim\sigma = k$ and refer to $\sigma$ as a \mbox{$k$-simplex}.
A simplicial complex $\mathcal{T}$ is called \emph{pure} or \emph{homogeneous} if all its maximal simplices have the same dimension~$d$.
Given a simplex $\sigma$, its \emph{link} is the subcomplex $\lk(\sigma) \coloneqq \{ \tau \in \mathcal{T} \mid \tau \cap \sigma = \emptyset, \tau \cup \sigma \in \mathcal{T} \}$.
If $\sigma = \{v\}$ is a \emph{vertex}, i.e., a \mbox{$0$-simplex}, the link can be seen as the boundary of a small neighborhood of~$v$ in the simplicial complex.
Given a \mbox{$d$-dimensional} homogeneous simplicial complex~$\mathcal{T}$, we say that $\mathcal{T}$ is a \emph{combinatorial \mbox{$d$-manifold}} if $\lk(v)$ is a triangulated \mbox{($d-1$)-sphere} for every vertex $v$.\footnote{
 For readers familiar with differential geometry/topology, this mimics the informal definition of a smooth $d$-manifold ``locally looking like'' $\mathbb{R}^d$.
}
While combinatorial manifolds are not restricted in terms of their dimension, we focus on $d \in \{2, 3\}$ because in this setting,
\emph{topological manifolds} coincide with \emph{combinatorial manifolds}~\citep{Moise77a}.
This has two implications, namely
\begin{inparaenum}[(i)]
    \item every combinatorial manifold \emph{triangulates} some topological manifold, and
    \item  the data augmentation procedures we subsequently introduce provide us with a way to represent \emph{all} topological manifolds in these dimensions as a combinatorial manifold, ensuring that our data is, in principle, a complete combinatorial census of low-dimensional manifolds.
\end{inparaenum}

\paragraph{Characterizing and representing triangulations.}
A topological manifold does not have \emph{one} unique combinatorial manifold, i.e., simplicial complex, assigned to it.
To measure whether two simplicial complexes triangulate the same manifold, we calculate \emph{topological invariants} like the \emph{Euler characteristic}~$\chi$. 
For a simplicial complex~$\mathcal{T}$, the Euler characteristic is the alternating sum of simplex counts in each dimension, denoted by
$\chi(\mathcal{T}) \coloneq \sum_{k=0}^d (-1)^k |\mathcal{T}_k|$.
Another invariant is the \emph{orientability} of a simplicial complex. 
We say that $\mathcal{T}$ is \emph{orientable} if we can assign \emph{consistent} orientations to all its \mbox{$d$-simplices}, meaning that for each \mbox{$(d-1)$-face} shared by two \mbox{$d$-simplices}, the induced orientations are \emph{opposite}.\footnote{
  Consider orienting all triangles of a triangulated surface by traversing them either clockwise or counterclockwise. The orientation is consistent if each edge is traversed once in each direction by its two adjacent triangles.
}
As we will describe below, the Euler characteristic and orientability are, taken together, \emph{sufficient} to fully characterize \mbox{$2$-manifolds}.
Finally, notice that a simplicial complex $\mathcal{T}$ admits several representations~(cf.\ \Cref{fig:triangulation-representation}).
The \emph{$1$-skeleton}~$\mathcal{S}$ of $\mathcal{T}$ is the graph formed by its vertices and edges; it typically loses a large amount of structural information.
The \emph{dual graph}~$\mathcal{D}$ of $\mathcal{T}$ has one vertex for each \mbox{$d$-simplex}, with two vertices connected by an edge if their corresponding simplices share a common \mbox{$(d-1)$}-simplex.
Last, the \emph{Hasse diagram}~$\mathcal{H}$ of $\mathcal{T}$ is the~(directed) graph on all simplices of $\mathcal{T}$, with an edge from $\sigma$ to $\tau$ whenever $\tau$ is a codimension-$1$ face of $\sigma$. \section{Topology-aware data augmentation and evaluation framework}

The richness of TDL models is in stark contrast to the lack of evaluation frameworks that contain \emph{inherently} higher-order objects. This scarcity is a limiting factor in the evaluation of new methods and models~\citep{Papamarkou24a, Rieck26a}. One of the few inherently higher-order datasets is the \texttt{MANTRA} dataset~\citep{ballester2025mantra}.  It contains triangulations of~(combinatorial) manifolds in $2$ and $3$ dimensions, originally collected by \citet{manifold_page}, and defines three tasks on them, namely
\begin{inparaenum}[(i)]
    \item Betti number prediction,
    \item homeomorphism type classification, and 
    \item orientability prediction.
\end{inparaenum}
Homeomorphism type classification fully characterizes
manifolds in all dimensions and constitutes one of the most fundamental tasks in
topology.
While this task is generally \emph{algorithmically undecidable} for dimensions $\ge 4$~\citep{Poonen14a}, classification theorems in dimensions $2$ and $3$ permit us to quantify the abilities of topological models in a controlled setting.\footnote{
    Readers interested in algorithmic aspects are referred to \citet[Chapter~9]{stillwell1993classical}.
}
However, one of the main \emph{limitations} of \texttt{MANTRA} is the lack of homeomorphism type labels. 
In the \mbox{$2$-dimensional} part of the dataset, only four manifold types were labeled, leaving $80\% $ of triangulations unlabeled.
The $3$-dimensional dataset has even more severe imbalances, with only three
classes and less than $0.5\%$ of the dataset having a different label than the
$3$-sphere.
Removing classes with few observations, as done by \citet{ballester2025mantra} for the $2$D task, severely restricts the \emph{diversity} of the dataset and, in $3$D, will result in a trivial classification task.
Hence, to derive meaningful results and investigate model properties required for manifold classification, we need to extend this dataset.

\subsection{Topological data augmentation}\label{subsection:topo_data_aug}

We first focus on how to remedy the lack of labels in $2$D by providing
information about the missing classes. The classification theorem of closed
surfaces~(\cref{theorem:class_2d}) and its corollary~(\cref{cor:inv_class})
permit us to classify \emph{all} triangulations of $2$D manifolds based on their Euler
characteristic and orientability.
To address the dataset \emph{imbalance}, we need to be able to create new triangulations with \emph{known} homeomorphism types. We thus employ \emph{Pachner moves} and \emph{connected sums}.
Pachner moves~\citep{Pachner1991} are \emph{local} transformations of simplicial complexes, which, if performed iteratively, can create \emph{all} possible triangulations of one manifold, i.e., one homeomorphism type.
\emph{Connected sums},\footnote{A standard operation that glues two manifolds along a shared boundary.
}
by contrast, change the homeomorphism type, albeit in a predictable way: Specifically, we remove a triangle from either a $T_2$~(torus) or an $\mathbb{RP}_2$~(projective plane) as well as from the original triangulation and glue the resulting spaces along their boundaries, using the classification theorem of $2$D manifolds to obtain a class label.
We thus obtain  three variations of the original \texttt{MANTRA} dataset:
\begin{itemize}[left = 0pt, itemsep = 0.125em]
    \item \textbf{$2$D-unbalanced}: We add class labels to all manifolds of \texttt{MANTRA} and  remove classes with fewer than $100$ samples, resulting in $9$ remaining classes.
    \item \textbf{$2$D-balanced}:
    Using Pachner moves and connected sums, we extend the number of observations in each class to $2,500$ samples.
The resulting dataset has $22$ classes and comprises $55,000$ observations, covering a significantly larger area of $2$D topology.
    \item \textbf{$3$D-balanced}: 
    We add additional triangulations of $3$D manifolds with small valence~\citep{manifold_page}. Similar to the $2$D dataset, we extend the number of observations per type until each class has $5,000$ samples. In this case, we only use Pachner moves, obtaining a dataset with $9$ classes.
\end{itemize}

\begin{figure}[tbp]
    \centering
    \subcaptionbox{Type $(1, 3) \leftrightarrow (3, 1)$\label{sfig:pachner-2d-1}}{\begin{tikzpicture}[font=\small,
  line join=round, line cap=round,
  vtx/.style = {circle, fill=black, inner sep=1.3pt},
  lbl/.style = {font=\scriptsize, inner sep=1pt},
  arr/.style = {-Stealth, thick, black!70},
]

  \def\xArrOff{2.0}

\begin{scope}[shift={(-\xArrOff,0)}]
    \coordinate (A) at (-0.85,-0.50);
    \coordinate (B) at ( 0.85,-0.50);
    \coordinate (C) at ( 0.00, 0.90);
    \fill[OIOrange!50] (A)--(B)--(C)--cycle;
    \draw[very thick, black!80] (A)--(B)--(C)--cycle;
    \node[lbl] at ($0.33*(A)+0.33*(B)+0.34*(C)$) {$\sigma_1$};
    \foreach \p in {A,B,C}\filldraw[black] (\p) circle (0.04);
    \node[lbl, below left]  at (A) {$v_1$};
    \node[lbl, below right] at (B) {$v_2$};
    \node[lbl, above]       at (C) {$v_3$};
  \end{scope}

\draw[arr] (-0.55, 0.10) -- ( 0.55, 0.10);
  \draw[arr] ( 0.55,-0.10) -- (-0.55,-0.10);
  \node[lbl, above] at (0, 0.10) {$(1,3)$};
  \node[lbl, below] at (0,-0.10) {$(3,1)$};

\begin{scope}[shift={(\xArrOff,0)}]
    \coordinate (A) at (-0.85,-0.50);
    \coordinate (B) at ( 0.85,-0.50);
    \coordinate (C) at ( 0.00, 0.90);
    \coordinate (D) at ( 0.00,-0.05);
    \fill[OISkyBlue!50]   (A)--(B)--(D)--cycle;
    \fill[OIGreen!50] (B)--(C)--(D)--cycle;
    \fill[OIYellow!50]  (C)--(A)--(D)--cycle;
    \draw[very thick, black!80] (A)--(B)--(C)--cycle;
    \draw[very thick, black!80] (A)--(D) (B)--(D) (C)--(D);
    \node[lbl] at ($0.33*(A)+0.33*(B)+0.34*(D)$) {$\sigma_2$};
    \node[lbl] at ($0.33*(B)+0.33*(C)+0.34*(D)$) {$\sigma_3$};
    \node[lbl] at ($0.33*(C)+0.33*(A)+0.34*(D)$) {$\sigma_4$};
    \foreach \p in {A,B,C,D}\filldraw[black] (\p) circle (0.04);
    \node[lbl, below left]  at (A) {$v_1$};
    \node[lbl, below right] at (B) {$v_2$};
    \node[lbl, above]       at (C) {$v_3$};
    \node[lbl, fill=white, inner sep=0.8pt, right=1pt] at (D) {$v_4$};
  \end{scope}

\end{tikzpicture} }
    \subcaptionbox{Type $(2, 2) \leftrightarrow (2, 2)$\label{sfig:pachner-2d-2}}{\begin{tikzpicture}[font=\small,
  line join=round, line cap=round,
  vtx/.style = {circle, fill=black, inner sep=1.3pt},
  lbl/.style = {font=\scriptsize, inner sep=1pt},
  arr/.style = {-Stealth, thick, black!70},
]

  \def\xArrOff{2.0}

\begin{scope}[shift={(-\xArrOff,0)}]
    \coordinate (A) at (-0.85, 0.00);
    \coordinate (B) at ( 0.00, 0.85);
    \coordinate (C) at ( 0.85, 0.00);
    \coordinate (D) at ( 0.00,-0.85);
    \fill[OIOrange!50]  (A)--(B)--(C)--cycle;
    \fill[OISkyBlue!50] (A)--(C)--(D)--cycle;
    \draw[very thick, black!80] (A)--(B)--(C)--(D)--cycle;
    \draw[very thick, black!80] (A)--(C);
    \node[lbl] at ($0.33*(A)+0.33*(B)+0.34*(C)$) {$\sigma_1$};
    \node[lbl] at ($0.33*(A)+0.33*(C)+0.34*(D)$) {$\sigma_2$};
    \foreach \p in {A,B,C,D}\filldraw[black] (\p) circle (0.04);
    \node[lbl, left]  at (A) {$v_1$};
    \node[lbl, above] at (B) {$v_2$};
    \node[lbl, right] at (C) {$v_3$};
    \node[lbl, below] at (D) {$v_4$};
  \end{scope}

\draw[arr] (-0.55, 0.10) -- ( 0.55, 0.10);
  \draw[arr] ( 0.55,-0.10) -- (-0.55,-0.10);
  \node[lbl, above] at (0, 0.10) {$(2,2)$};
  \node[lbl, below] at (0,-0.10) {$(2,2)$};

\begin{scope}[shift={(\xArrOff,0)}]
    \coordinate (A) at (-0.85, 0.00);
    \coordinate (B) at ( 0.00, 0.85);
    \coordinate (C) at ( 0.85, 0.00);
    \coordinate (D) at ( 0.00,-0.85);
    \fill[OIGreen!50]  (A)--(B)--(D)--cycle;
    \fill[OIYellow!50] (B)--(C)--(D)--cycle;
    \draw[very thick, black!80] (A)--(B)--(C)--(D)--cycle;
    \draw[very thick, black!80] (B)--(D);
    \node[lbl] at ($0.33*(A)+0.33*(B)+0.34*(D)$) {$\sigma_3$};
    \node[lbl] at ($0.33*(B)+0.33*(C)+0.34*(D)$) {$\sigma_4$};
    \foreach \p in {A,B,C,D}\filldraw[black] (\p) circle (0.04);
    \node[lbl, left]  at (A) {$v_1$};
    \node[lbl, above] at (B) {$v_2$};
    \node[lbl, right] at (C) {$v_3$};
    \node[lbl, below] at (D) {$v_4$};
  \end{scope}
\end{tikzpicture} }
    \caption{Pachner moves on the triangulation of a $2$-manifold. Each move~(and its inverse) constitutes a local re-triangulation that does not change the underlying topological type of the manifold.
    }
    \label{fig:pachner}
\end{figure}

\paragraph{Deduplication and data leakage prevention.}
Applying Pachner moves~(cf.\ \Cref{fig:pachner}) and connected sums can result in isomorphic triangulations, causing duplicates in the dataset. To prevent this and avoid leakage from training into test data, we first oversample the required triangulations by a factor of two. Afterwards, we apply the following set of heuristics:
\begin{enumerate}[left = 0pt, itemsep = 0.125em]
    \item We compute the $f$-vector, i.e., the vector containing the number of $i$-dimensional simplices at each index $f_i$. Simplicial complexes that differ in their $f$-vector cannot be isomorphic.
    \item We compute the \emph{incidence graph}, i.e., a graph consisting of the face and vertices as nodes and edges between a face-node and a vertex-node pair, if the node is included in the face. Iteratively, we then calculate the $f$-vector and the WL-hash of the incidence graph~\citep{Morris23a}. Triangulations whose incidence graphs can be distinguished by this filtering are guaranteed to be non-isomorphic.
    \item Finally, we perform direct isomorphism checks  on the incidence graph. If the remaining subsets are too large, every triangulation but one per subset gets removed.
\end{enumerate}
We finally ensure that no triangulation with more than $20$ vertices for $2$D and $30$ vertices for $3$D is present to allow the creation of unseen test data with more vertices. Generating and deduplicating the full balanced datasets takes \qty{\approx 1.1}{\minute}~($2$D, \SI{0.8}{\giga\byte} peak RAM) and \qty{\approx 14.5}{\minute}~($3$D, \SI{1.7}{\giga\byte} peak RAM) on a single MacBook Pro M4 core.

\begin{tcolorbox}
    We extend the \texttt{MANTRA} dataset to a more \emph{varied} set of manifolds in both $2$D and $3$D through the inclusion of previously-collected triangulations and triangulations obtained via Pachner moves and connected sums. This procedure is \emph{not} limited to this dataset, but can be applied to all data that affords a simplicial complex representation and is invariant with respect to homeomorphism. 
\end{tcolorbox}

\subsection{Quantifying topology means quantifying topological generalization}
\label{sec:topological-generalization}

The primary objective of this benchmark is to provide a principled evaluation to quantify whether an architecture, with a given set of inductive biases, is learning \emph{topological} properties of the input data or just exploiting \emph{combinatorial} artifacts.
Before introducing our novel evaluation methodology, we unpack key traits that we expect of models that claim to learn topological information from data.

\paragraph{Topology versus combinatorics.}
An important distinction between topology and combinatorics lies in the objects they aim to distinguish and under which operations these objects are considered to be ``the same.''
Topology is interested in distinguishing objects~(like combinatorial manifolds) up to homeomorphism. That is, two manifolds are considered the same~(homeomorphic) if there exists 
a bijective continuous function between them whose inverse is continuous.
For combinatorial manifolds, the natural notion of equivalence is \emph{PL homeomorphism}, i.e., two triangulations are considered to be the same if they are connected by a finite sequence of Pachner moves.
Combinatorics, by contrast, studies the \emph{combinatorial structure} of complexes~\citep{Hajij23a}, i.e., properties invariant under vertex relabeling but \emph{not} necessarily under homeomorphism.
Hence, two simplicial complexes are considered to be the same~(\emph{isomorphic}), if one can be obtained from the other by vertex relabeling.
We believe that these two notions of equivalence are inadvertently conflated, leading models to pick up \emph{triangulation-specific artifacts} rather than \emph{topological invariants}.
While this can and does lead to high predictive performance in tasks where
topology is potentially not required to solve a task~\citep{Horn22a}, there are
also application domains for which topology-aware inductive biases are
indispensable, such as drug design~\citep{Cang17a}, materials
science~\citep{Minamitani25a}, or physics~\citep{Cole19a, Donato16a, Xu19a}.
We find that these areas are not fully represented in existing benchmarks,
meaning that models may be unintentionally optimized for the wrong objectives and fail
silently despite a topological bias being necessary.

\paragraph{Exploring different scales of topology.}
To quantify to what extent a model is focusing on a combinatorial or a
topological signal, we need operations that change simplicial complexes without
changing the underlying manifold.
A \emph{subdivision} or \emph{refinement} is a modification of a simplicial
complex that adds additional simplices while preserving topological
properties.
Refinements are combinations of Pachner moves that can increase the number of
vertices, thus leading to a change of scale. For example,
\Cref{sfig:pachner-2d-1} depicts a Pachner move that is also a subdivision,
since $\sigma_1$ was replaced with three new triangles, i.e.,
$\sigma_2,\sigma_3,\sigma_4$.
For a triangulation $\mathcal{T}$, we define two important types of
subdivisions that we will subsequently use.
A \emph{stellar subdivision} of a maximal simplex $\sigma \in \mathcal{T}$ is obtained by
\begin{inparaenum}[(i)]
    \item adding a new vertex $v$,
    \item removing the simplex $\sigma$, and
    \item adding the simplices $\tau \cup \{v\}$ for every face $\tau \subsetneq \sigma$ to $\mathcal{T}$.
\end{inparaenum}
This operation results in \emph{one} additional vertex.
\begin{wrapfigure}[11]{r}{0.3\linewidth}
    \vspace{-\baselineskip}
   \centering
   \begin{tikzpicture}[scale=1.5,font=\small,
  line join=round, line cap=round,
  vtx/.style = {circle, fill=black, inner sep=1pt},
  lbl/.style = {font=\scriptsize, inner sep=1pt},
]
  \coordinate (A) at (-0.85,-0.50);  \coordinate (B) at ( 0.85,-0.50);  \coordinate (C) at ( 0.00, 0.90);  \coordinate (E) at ($0.5*(A)+0.5*(B)$);  \coordinate (F) at ($0.5*(B)+0.5*(C)$);  \coordinate (G) at ($0.5*(A)+0.5*(C)$);  \coordinate (D) at ($0.333*(A)+0.333*(B)+0.334*(C)$);  

\fill[OIOrange!50]  (A)--(E)--(D)--cycle;
  \fill[OISkyBlue!50] (E)--(B)--(D)--cycle;
  \fill[OIOrange!50]   (B)--(F)--(D)--cycle;
  \fill[OISkyBlue!50]  (F)--(C)--(D)--cycle;
  \fill[OIOrange!40]  (C)--(G)--(D)--cycle;
  \fill[OISkyBlue!45]     (G)--(A)--(D)--cycle;

\draw[very thick, black!80] (A)--(B)--(C)--cycle;
\foreach \q in {A,B,C,E,F,G}
    \draw[thick, black!70] (D)--(\q);

\node[lbl] at ($0.33*(A)+0.33*(E)+0.34*(D)$) {$\sigma_2$};
  \node[lbl] at ($0.33*(E)+0.33*(B)+0.34*(D)$) {$\sigma_3$};
  \node[lbl] at ($0.33*(B)+0.33*(F)+0.34*(D)$) {$\sigma_4$};
  \node[lbl] at ($0.33*(F)+0.33*(C)+0.34*(D)$) {$\sigma_5$};
  \node[lbl] at ($0.33*(C)+0.33*(G)+0.34*(D)$) {$\sigma_6$};
  \node[lbl] at ($0.33*(G)+0.33*(A)+0.34*(D)$) {$\sigma_7$};

\foreach \p in {A,B,C,D,E,F,G}
    \filldraw[black] (\p) circle (0.03);

\node[lbl, below left]                             at (A) {$v_1$};
  \node[lbl, below right]                            at (B) {$v_2$};
  \node[lbl, above, yshift = 2pt]                    at (C) {$v_3$};
  \node[lbl, fill=white, inner sep=0.8pt, right=1pt] at (D) {$v_4$};
  \node[lbl, below, yshift = -2pt]                   at (E) {$v_5$};
  \node[lbl, right=1pt]                              at (F) {$v_6$};
  \node[lbl, left=1pt]                               at (G) {$v_7$};
\end{tikzpicture}
    \caption{Barycentric subdivision of a triangle $\sigma_1$ into 6 triangles ($\sigma_2$ to $\sigma_7$).}
   \label{fig:full_bary}
\end{wrapfigure}
A \emph{barycentric subdivision}~\citep[§15]{MR755006}
denotes a triangulation $\mathcal{T}'$ whose $d$-dimensional faces are sequences of strict inclusions of simplices of $\mathcal{T}$~(cf.\ \Cref{fig:full_bary}).
A barycentric subdivision not only adds triangles and vertices but also intermediate simplices like edges.
Applying any number of subdivisions results in fixed combinatorial changes to the underlying simplicial complex without changing its homeomorphism type.
We exploit the topological invariance under subdivisions to extend our benchmark
and stress-test models.
By keeping \emph{topology} fixed while varying the \emph{combinatorial}
structure, we induce an incremental shift leading outside of the original combinatorial
training distribution.
That  way, we can evaluate whether the models are learning topological
properties and at which point they start to fail to generalize.
For each of the aforementioned versions of the data, we introduce a set of
additional variations based on subdivisions to test already trained models.
Due to the growth of the complex induced by incremental subdivision, we restrict
these additional test datasets to $100$ samples per type of
triangulation, yielding the following variants:
\begin{itemize}[left = 0pt, itemsep = 0.125em]
    \item \textbf{$n$-graded stellar subdivision}. We perform stellar subdivisions on the maximal simplices iteratively until we reach a simplicial complex with $n$ vertices. To ensure that there are no repeated triangulations, we require $n$ to be larger than the number of vertices in the largest triangulation.
    
    \item \textbf{$p$-top stellar subdivision}. We perform one step of stellar subdivision on a proportion $p \in \{0.75, 1\}$ of the maximal simplices of $\mathcal{T}$.
    
    \item \textbf{Barycentric subdivision}. We perform a \emph{full} barycentric subdivision, that is, we perform a barycentric subdivision of all simplices of the simplicial complex. For this, we pick the $100$ largest triangulations of each class in the reference dataset.
\end{itemize}
The maximum number of vertices for $2$D-unbalanced, $2$D-balanced, and
$3$D-balanced are $15$, $20$ and $30$, respectively. For each, we create the
graded subdivisions $n \in \{16, \dots, 20\}$, $n \in \{21, \dots,
24\}$ and $n \in \{31, \dots, 34\}$ accordingly.
Additionally, we add the three variations ($0.75$-top stellar, $1$-top stellar, and
barycentric).
Next, we use these datasets to assess how different models capture topology. 

\begin{tcolorbox}
    We propose novel data augmentation methods to allow researchers of topological models to benchmark their architectures on a \emph{controlled} combinatorial out-of-distribution setting that keeps topological properties fixed. Our methodology permits controlled, incremental refinement and applies to all triangulations, while providing a \emph{robust assessment} of predictive models.
\end{tcolorbox} \section{Are we learning combinatorics or topology?}

Using the three training datasets and model evaluation pipeline with subdivision evaluation datasets presented in the previous section, we are now presenting our experimental setup and results. 

\subsection{Experimental setup}

\begin{wraptable}{r}{0.4\linewidth}
    \vspace{-\baselineskip}
    \centering
    \caption{The symbols denote the degree of attainment of each \emph{criterion}. The $\mathcolor{GHigh}{\checkmark}$ is complete,  $\mathcolor{OISkyBlue}{\sim}$ is partial and $\mathcolor{OIBlue}{\times}$ is lack. The $\dagger$ marks the variations we choose.
    }
    \label{tbl:eval_models}
      \begin{tabular}{lcccc}
        \toprule
         Model & \textbf{R} & \textbf{A} & \textbf{V} & \textbf{E}  \\
        \midrule
         SCCNN $(\dagger)$ &  $\mathcolor{OISkyBlue}{\sim}$ &$\mathcolor{GHigh}{\checkmark}$  & $\mathcolor{OISkyBlue}{\sim}$ & $\mathcolor{GHigh}{\checkmark}$\\
        CWN $(\dagger)$  & $\mathcolor{OISkyBlue}{\sim}$ & $\mathcolor{GHigh}{\checkmark}$& $\mathcolor{OIBlue}{\times}$ & $\mathcolor{GHigh}{\checkmark}$\\
        GCCN  & $\mathcolor{GHigh}{\checkmark}$ & $\mathcolor{GHigh}{\checkmark}$ & $\mathcolor{GHigh}{\checkmark}$ & $\mathcolor{OIBlue}{\times}$\\
        SMCN  & $\mathcolor{OISkyBlue}{\sim}$ & $\mathcolor{GHigh}{\checkmark}$ & $\mathcolor{GHigh}{\checkmark}$ & $\mathcolor{OIBlue}{\times}$ \\
        CT  & $\mathcolor{GHigh}{\checkmark}$ & $\mathcolor{OIBlue}{\times}$ & $\mathcolor{OIBlue}{\times}$ & $\mathcolor{OIBlue}{\times}$ \\
        \bottomrule
      \end{tabular}
\end{wraptable}
We compare representations and encodings of triangulations on our two target domains, i.e., graphs and simplicial complexes, selecting models by increasing expressivity based on four criteria:
\begin{inparaenum}[(i)]
    \item \textbf{R}elevance as a measure of applicability to the data domain,
    \item \textbf{A}vailability of the implementation,
    \item \textbf{V}erifiability of the theoretical proposal with respect to the implementation, and
    \item computational \textbf{E}fficiency of executing the model.
\end{inparaenum}
For graphs, we pick GCN~\citep{kipf2017semisupervised}, Residual Gated GCN~\citep[RG-GCN]{bresson2017residual}, and Graphormer~\citep{ying2021graphormer}.
For simplicial complexes, \emph{no} model satisfies all criteria~(cf.\ \Cref{tbl:eval_models,tbl:full_eval_models}):
Simplicial Convolutional Neural
Networks~\citep[SCNN]{Yang2022simplicial} required non-trivial modifications to
the published code.
Furthermore, at least two different implementations of Simplicial Complex Convolutional Neural
Networks~\citep[SCCNN]{yang2025hodgeaware} exist, described in  \texttt{TopoX}~\citep{topox} and \texttt{TopoBench}~\citep{telyatnikov2025topobench}, respectively, both of which necessitate substantial changes.
As for additional models, CIN~\citep{bodnar2021cwnetworks} and CXN~\citep{hajij2020cell} operate on cell complexes. CIN's implementation is memory-inefficient while CXN provides none. \texttt{TopoBench} provides an implementation of CWN, which, however, differs substantially from its original description~\citep{bodnar2021cwnetworks}.
Moreover, the attention-based SAN~\citep{goh2022simplicial} has a defunct, computationally-inefficient implementation, whereas CT~\citep{barsbey2025higherorder, Ballester24a} has none.
Approaches based on the \emph{augmented Hasse diagram}~\citep[p.\ 45]{Hajij23b}, such as GCCN~\citep{papillon2025topotune} and SMCN~\citep{eitan2025topological} are \emph{infeasible} to run on datasets beyond the small \texttt{TUDatasets}~\citep{Morris+2020}.
Apart from GCCN and SMCN, \emph{no model} operates on complexes of dimension higher than $2$.
Simple extensions of SCCNN and CWN to use \mbox{$3$-simplices} raise the runtime significantly, making it infeasible to train. Due to a range of conflicting implementations, we decided to keep modifications minimal. Thus, SCCNN accounts for $3$-simplices only through connectivity via the Hodge Laplacian and does not carry features for them, whereas CWN is limited to $2$-simplices.
We believe that the lack of publicly-available implementations capable of handling $3$D triangulations at scale poses a serious bottleneck for advancing TDL.
Finally, we did not include models based on the Euler characteristic transform like DECT~\citep{Roell24a} since the Euler characteristic is a strong predictor on this dataset.

\paragraph{Representations.}
The canonical representation of a triangulation is an~(abstract) simplicial complex~$(\mathcal{T})$. When storing full boundary matrices\footnote{Storing top-level simplices is sufficient for a manifold, but there are at present no TDL models making use of this shortcut.
} in GPU memory, this representation can become impracticable.
As an alternative, we explore representations with varying degrees of information preservation. We then evaluate whether such parsimonious representations still capture the desired traits. Subsequently, we use the $1$-skeleton $(\mathcal{S})$, the dual graph $(\mathcal{D})$, and the Hasse diagram $(\mathcal{H})$ as representations.
To ensure that the training remains unbiased and cannot exploit combinatorial artifacts, we randomly permute all vertices of the triangulation before processing it.

\paragraph{Computational complexity.}
\Cref{tbl:comp_graph,tbl:comp_sc} show the computational complexity of HOMP and
graph models. The computational complexity of message-passing on the Hasse
diagram is lower than for any of the HOMP methods, while the Graphormer becomes
infeasible to run on the Hasse diagram for $3$D data due to its quadratic
scaling. 

\paragraph{Encodings.} We also investigate the role of \emph{feature encodings} in conjunction with different representations and architectures. To do so, we define the encoding of a representation as the set of \emph{features} assigned to its vertices or simplices, respectively, denoting both assignments with $x$ for notational simplicity. Subsequently, we use four encoding methods:
\begin{enumerate}[left = 0pt, itemsep = 0.125em]
    \item Random uniform features ($\mathbf{R}$) in $[0,1]^k$, with $k$ being a tunable hyperparameter~(cf.\ \Cref{section:app_hyperparams});
\item Node Degree ($\mathbf{D}$);
\item Random Walk Positional Encoding ($\mathbf{RWPE}$, \citealp{dwivedi2022graph});
    \item Moment Curve Embedding ($\mathbf{MC}$): For a \mbox{$d$-dimensional} triangulation and $x$ with index $i$, we set $x=\left[\left(\frac{i}{n-1}\right)^1, \dots,\left(\frac{i}{n-1}\right)^{2d+1}\right]$.
\end{enumerate}
For graph representations, we use all encodings, whereas for simplicial
complexes, we only apply $\mathbf{R}$ and $\mathbf{MC}$, since they are
well-defined on both domains. For simplices, we calculate the
$\mathbf{MC}$ encoding based on the barycenter of the embedding
of \mbox{$0$-simplices}. 

\paragraph{Evaluation procedure.}
First, we randomly split our dataset into 60\% training data, 20\% validation data, and 20\% test data. Then, we train each model over a set of hyperparameters, performing a grid search. Similar to graph benchmarking papers for this dataset size~\citep{dwivedi2022graph}, we restrict our parameter budget to a maximum of $500$K for each model to ensure fair comparisons. We select the hyperparameter configuration with the best average validation data performance over three seeds and evaluate this configuration on the held-out test set. We train all models for up to $300$ epochs, with an early stopping of $100$ epochs; we empirically observed that either the losses stabilize or the training saturates in these ranges.
\Cref{app:hardware_config} provides the hardware and software configuration for the training setup, while \Cref{section:app_hyperparams} lists  the full hyperparameter configurations for each model.

\begin{table}[btp]
\centering
\caption{Best representation/encoding per model on $2$D triangulations for the
    unbalanced and balanced settings. $\Delta_T$ denotes the difference to the theoretical
    distinction capability based on the \emph{Euler characteristic}.
}
\label{tab:results-2d}

\sisetup{detect-all=true,detect-weight=true,table-align-uncertainty=true,table-comparator=true,table-format=<2.2(1.2),tight-spacing=true}

\resizebox{\linewidth}{!}{\begin{tabular}{l l lSc lSc}
\toprule
 & & \multicolumn{3}{c}{\textbf{Unbalanced}} & \multicolumn{3}{c}{\textbf{Balanced}} \\
\cmidrule(lr){3-5}\cmidrule(lr){6-8}
Type & Model & Repr.\,/\,Enc. & \text{Bal.\ Acc.\ $(\uparrow)$} & $\Delta_T$ & Repr.\,/\,Enc. & \text{Bal.\ Acc.\ $(\uparrow)$} & $\Delta_T$ \\
\midrule
\multirow{3}{*}{\textbf{GNNs}}
& GCN & $\mathcal{D}$ / RWPE & 84.12 (0.24) & \textcolor{GLow}{+\num{6.32}} & $\mathcal{D}$ / RWPE & 89.55 (0.48) & \textcolor{GMid}{+\num{21.37}} \\
& RG-GCN & $\mathcal{D}$ / RWPE & 87.35 (0.44) & \textcolor{GMid}{+\num{9.55}} & $\mathcal{D}$ / RWPE & 92.30 (0.46) & \textcolor{GHigh}{+\num{24.12}} \\
& Graphormer & $\mathcal{H}$ / RWPE & 94.45 (2.56) & \textcolor{GHigh}{+\num{16.65}} & $\mathcal{H}$ / RWPE & 96.85 (0.31) & \textcolor{GHigh}{+\num{28.67}} \\
\midrule
\multirow{2}{*}{\textbf{HOMP}}
& SCCNN & $\mathcal{T}$ / R & 86.50 (7.51) & \textcolor{GMid}{+\num{8.70}} & $\mathcal{T}$ / MC & 87.93 (0.31) & \textcolor{GMid}{+\num{19.75}} \\
& CWN & $\mathcal{T}$ / MC & 77.53 (0.05) & \textcolor{GLow}{-\num{0.27}} & $\mathcal{T}$ / MC & 71.10 (1.77) & \textcolor{GLow}{+\num{2.92}} \\
\bottomrule
\end{tabular}
}

\end{table} 
\subsection{Analyzing the impact of representations and encodings}

Our goal is to investigate the ability of current GNN and TDL models to differentiate manifolds, based on triangulations that contain \emph{all} information required to distinguish them.
We observe that the performance for $2$D triangulations is varied, with GNNs showing clear advantages over HOMP models, whereas the differences are less pronounced in the $3$D case.
\Cref{tab:results-2d} shows the performance of the best combination of encoding and representation per model for the $2$D classification task.
We show both the unbalanced and balanced versions side by side, as well as the best choice of representation and encoding, respectively.
Overall, the $2$D datasets reveal a hierarchy of model architectures,
with GNNs \emph{consistently} outperforming higher-order methods.
Notably, SCCNN exhibits training instability for the unbalanced dataset.
The balancing in $2$D increases the difficulty of the task by
introducing more classes while presenting the model with a larger variety of topological structures. This appears to be particularly beneficial for GNNs, as they outperform the HOMP
architectures. Interestingly, GNN performance seems to be correlated with expressivity, i.e., more expressive architectures exhibit higher predictive performance. CWN performance, by contrast, remains comparable to the Euler
characteristic.
This is noteworthy since it contradicts prior
work~\citep{ballester2025mantra}; our results imply that less computationally
complex models like GCN and RG-GCN can solve the task sufficiently well
using the dual graph, and may thus be ultimately
preferable to HOMP models.

\begin{wraptable}{l}{0.50\linewidth}
    \centering
    \sisetup{
    detect-all=true,detect-weight=true,table-align-uncertainty=true,table-comparator=true,table-format=2.2(1.2),tight-spacing=true,uncertainty-mode = separate,
    }
    \small
    \caption{Best representation/encoding per model on $3$D triangulations. $\Delta_T$ denotes the difference in performance w.r.t. a random forest model trained on the $f$-vector.
    }
    \label{tab:results-3d}
\begin{tabular}{@{}llSr}
        \toprule
         Model & Repr.\,/\,Enc. & {Bal.\ Acc.\ $(\uparrow)$} & $\Delta_T$ \\
        \midrule
        GCN & $\mathcal{H}$ / MC & 89.83(0.08) & \textcolor{GHigh}{+\num{7.39}} \\
        RG-GCN & $\mathcal{H}$ / RWPE & 89.82(0.15) & \textcolor{GHigh}{+\num{7.38}} \\        Graphormer & $\mathcal{D}$ / RWPE & 89.57(0.42) & \textcolor{GHigh}{+\num{7.13}} \\
        \midrule
        SCCNN & $\mathcal{T}$ / R & 82.69(0.46) & \textcolor{GLow}{+\num{0.25}} \\
        CWN$^*$ & $\mathcal{T}$ / MC & 84.60(0.64) & \textcolor{GMid}{+\num{2.16}} \\
        \bottomrule
        \end{tabular}
    \flushleft\footnotesize$^*$: Up to 2-simplices only
\end{wraptable} 
\cref{tab:results-3d} shows the performance on triangulations of $3$D manifolds. They all share the \emph{same} Euler characteristic, so the baseline comparison is a random forest model trained on the $f$-vector.
In this case, the differences in predictive performance are less pronounced. 
All graph models perform on a similar level, with non-attention models
performing best on the Hasse diagram and attention-based ones best on the dual
graph, thus showing that model capacity is not a strict limitation, provided the right combination of representation and encoding is used.
Surprisingly,  once again, the performance of HOMP models lags behind graph
models. Notice, however, that CWN in its current implementation lacks the
capability to use \mbox{$3$-simplices}, meaning it is restricted to triangles,
edges, and vertices. Interestingly, SCCNN is outperformed by all GNNs.

\subsection{Current architectures appear to learn combinatorics}

\begin{figure}[tbp]
    \centering
    \includegraphics[width=0.95\linewidth,]{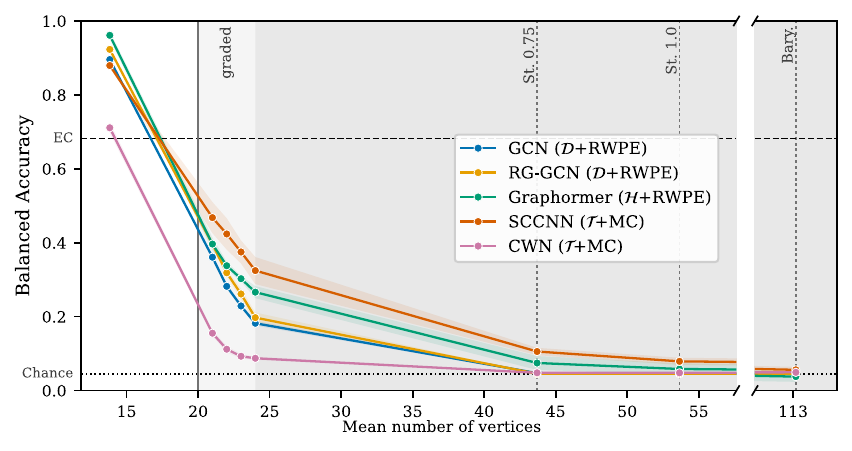}
    \caption{Balanced accuracy on the $2$D-balanced data under increasingly fine subdivisions. The leftmost data point denotes the in-distribution test performance while the shaded regions cover the out-of-distribution subdivision experiments. Balanced accuracy drops below the Euler characteristic heuristic~(EC, dashed) after a \emph{single} graded subdivision and decays to chance level~(dotted).
    }
    \label{fig:ood_2d_bal}
\end{figure}

\begin{figure}[tbp]
    \centering
    \includegraphics[width=0.95\linewidth,]{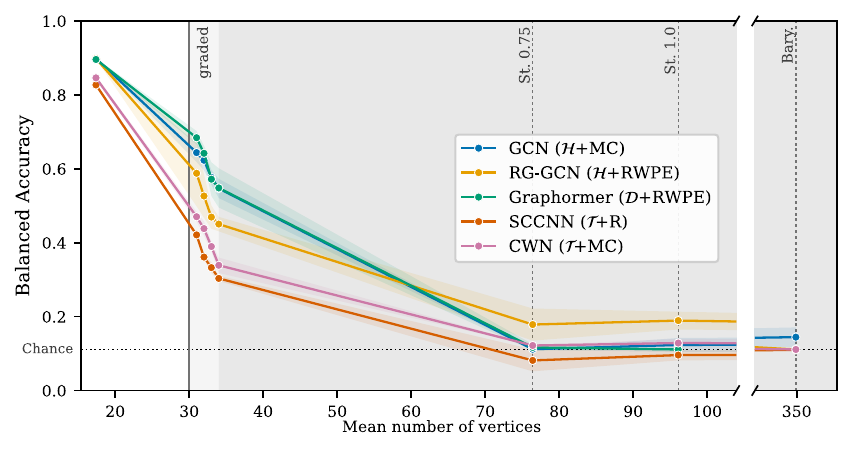}
    \caption{Balanced accuracy for subdivisions of the $3$D-balanced data.
    The leftmost data point denotes the in-distribution test performance while the shaded regions cover the out-of-distribution subdivision experiments
With increasing subdivision steps, performance decreases until chance level.
    }
    \label{fig:ood_3d_bal}
\end{figure}

All architectures can at least partially learn the combinatorial structure of the dataset, provided that they have access to an \emph{appropriate} representation and encoding. 
We now assess whether this implies that they are also learning the topology of the data.
The main objective of our work is to provide evaluation techniques and recommendations, and the results above demonstrate why we should be more critical of current architectures. The aim of topological deep learning is not only to develop models that are aware of combinatorial structure but also models that are aware of topological information: The data we are dealing with are combinatorial representations of continuous manifolds, so the object to characterize is the underlying manifold and \emph{not} any particular representation of it.
To assess the capability of a model to accomplish this, we now measure
performance on the subdivision datasets~(see \Cref{sec:topological-generalization}), evaluating 
 models on increasingly-refined subdivisions to gauge topological generalization.
\Cref{fig:ood_2d_bal,fig:ood_3d_bal} show the 
 predictive performance of the \emph{best} representation and encoding on the balanced datasets, based on lowest in-distribution validation loss, while \Cref{fig:ood_2d_ubal} depicts the performance for the unbalanced data.
The first data point in each figure represents in-distribution balanced accuracy in the classification task.
The remaining data points represent~(in order) the out-of-distribution datasets, i.e., $n$-graded stellar (\emph{graded}), $0.75$-top stellar (\emph{St.}\ $0.75$), $1$-top stellar (\emph{St.}\ $1.0$), and barycentric (\emph{Bary.}) subdivision, respectively.
For all datasets and models, we observe a \emph{sharp} decrease in performance after the first graded subdivision. 
In $2$D, the performance even falls below a deterministic heuristic based on the Euler characteristic~(notice that in $3$D, the manifolds are indistinguishable by the Euler characteristic alone, so we do not report its performance in the plot).
As the number of subdivision steps increases, all models eventually reach
chance-level performance.
Surprisingly, the performance drop is present in the
$2$D-unbalanced, $2$D-balanced, and $3$D-balanced scenarios. Since subdivisions
are merely a sequence of Pachner moves, the addition of Pachner-augmented samples to the training set would be expected to improve performance.
The decay in performance suggests two insights, namely
\begin{inparaenum}[(i)]
    \item the need for better evaluation methods for architectures that \emph{claim} to learn topology, addressed in this benchmark, and 
    \item  the need for new architectures that \emph{actually} learn topology. 
\end{inparaenum}
We argue that the community has moved prematurely to the second step, and we believe that our evaluation helps in addressing this gap.

\begin{tcolorbox}
  In $2$ and $3$ dimensions, models fail to generalize beyond the combinatorial structure, \emph{regardless
  of representation and encoding}, and appear to be unable to learn inherent
  topological properties, even on simple low-dimensional test cases.
\end{tcolorbox}

\section{Discussion}

Our experiments have immediate consequences for evaluating models and developing future models.
In terms of \emph{evaluations}, we believe that a dataset like \texttt{MANTRA} should be considered a ``smoke test,'' i.e., a topological equivalent of the \texttt{MNIST} dataset.
This is due to the fact that the classification of $2$- and \mbox{$3$-manifolds} is something that every model that claims to learn topological structure should be capable of doing: The targets are well understood and known to be computable from basic invariants like the Euler characteristic, orientability, and torsion. However, \emph{computable} does not imply \emph{efficiently computable}; for instance in the $3$D case the complexity is exponential~\citep{lackenby2020algorithms}. We thus posit that any model that fails to saturate this benchmark cannot be considered topological.

\paragraph{Recommendations.}
Due to the heavily-skewed class distribution in the original \texttt{MANTRA} dataset, we strongly recommend users to apply our proposed \emph{data augmentations}, i.e., Pachner moves and connected-sum gluings. This results in a more nuanced evaluation and ensures balanced classes. Notably, this recommendation is general and applies to any future datasets of a topological nature: Since our augmentations leave the underlying homeomorphism type unchanged, they can be used to increase dataset diversity and afford fair comparisons.
Finally, to assess \emph{generalization performance}, we observe that standard random train--val--test splits in \texttt{MANTRA} and other datasets are only providing a combinatorial rather than a topological evaluation due to model capacity. Our proposed \emph{refinement strategy} permits disentangling these two aspects while avoiding data leakage. Thus, we recommend that practitioners adopt these strategies to test to what extent a model is capable of picking up a topological signal~(resulting in sustained predictive performance across refinement steps) as opposed to a combinatorial one~(resulting in deteriorated predictive performance across refinement steps).

\paragraph{Lessons learned.}
We draw additional lessons from our experiments, the first one being related to \emph{computational aspects}. Specifically, we observe a large gap between the theoretical claims of models operating on simplicial complexes and what can be practically validated. This problem is largely due to scaling issues, and, echoing recent recommendations on improving graph learning~\citep{Bechler-Speicher25a, Coupette25a}, we strongly recommend that future evaluations \emph{discard} small graph datasets that employ ad-hoc lifting techniques in favor of inherently higher-order datasets~\citep{Rieck26a}.
This requires a shift in benchmarking culture and a conscious effort to move \emph{beyond} combinatorial expressivity, which is not the correct perspective to assess the topological diversity that we want to learn and that models purport to learn.
Adding to this, our central experimental finding is that \emph{all} considered models \emph{fail to generalize} even in the regime of modest refinements. This is an alarming observation since it strongly suggests that the models, having sufficient capacity to do so, exploit combinatorial artifacts to achieve high predictive performance on small triangulations without learning \emph{any} topological information.
In that setting, models are even incapable of matching the Euler characteristic, an invariant they are all theoretically capable of computing, let alone slightly more complex properties like orientability. This complements the findings by \citet{eitan2025topological}, who point out that HOMP models cannot learn certain invariants---whereas our experiments demonstrate that they do not even learn the ones that they are theoretically capable of learning.
We thus also provide novel insights into the structure-versus-features debate originating in graph learning~\citep{Bechler-Speicher25a, Coupette25a}. 
Specifically, we observe that it is possible to saturate the benchmark with
certain architectures, such as Graphormer on $2$D-balanced, and
different architectures require different configurations to achieve equivalent
performance.
However, we were unable to detect a clear pattern concerning the representation and features that would consistently result in high predictive performance.

\paragraph{Future work.}
Nevertheless, we can state that GNNs, in line with recent claims~\citep{peixoto2026graphsmaximallyexpressivehigherorder}, do not seem to exhibit an \emph{a priori} disadvantage when it comes to expressivity in capturing higher-order interactions.
While high predictive performance remains contingent on a suitable choice of representation and features, we find that a random walk positional encoding can recover some of the topological information discarded when creating a graph representation.
Notably, HOMP models and GCN in $3$D appear to benefit more from a moment curve embedding despite having access to the \emph{full} simplicial complex so that the moment
curve does not add any new information. We consider the analysis of this phenomenon to be an interesting direction of future research.

\begin{tcolorbox}
The playing field between GNNs and HOMP models is thus somewhat level again, with HOMP models not exhibiting favorable performance on higher-order datasets, but unfortunately at the cost of having \emph{no} model that is capable of addressing even simple tasks in low-dimensional topology.
We believe that the topological structure of data is a rich source of information that models have yet to fully leverage, and we therefore hope that our evaluation framework will serve as a stepping stone towards models that are not only \emph{inspired} by topological concepts but genuinely \emph{informed} by them.
\end{tcolorbox} 
\begin{ack}
This work has received funding from the Swiss State Secretariat for
Education, Research, and Innovation~(SERI).
This research was funded by the Swiss National Science Foundation~(SNSF) via grant IZSEZ0\_235713.
G.\,W.\ was partially funded by Humboldt Research Fellowship, CIFAR AI Chair, NSERC Discovery grant 03267, FRQNT grant 343567, and NSF grant DMS-2327211.
The authors declare no competing interests. The funders had no role in
the preparation of the manuscript or the decision to publish.
\end{ack}

\bibliographystyle{abbrvnat}
\bibliography{biblio}

\clearpage

\appendix
\crefalias{section}{appendix}

\startcontents
\printcontents{}{1}{{\vskip10pt\hrule
    \large\textbf{Appendix~(Supplementary Materials)}\vskip3pt\hrule\vskip5pt}
}

\counterwithin*{figure}{part}
\stepcounter{part}
\renewcommand{\thefigure}{S.\arabic{figure}}

\counterwithin*{table}{part}
\stepcounter{part}
\renewcommand{\thetable}{S.\arabic{table}}

\section{Changelog}

We provide a brief list of changes that arose in the time between the review
period and the camera-ready version.

\begin{compactitem}
    \item Changes in the \emph{main text}:
        \begin{compactitem}
            \item We perform deduplication now by first oversampling followed by the previous deduplication; this improves computational performance and we added the respective timings.
            \item Consequently, we managed to generate more samples for underrepresented classes and re-ran all experiments.
            \item We fixed the tables to show \emph{balanced accuracy} instead of \emph{standard accuracy}; the originally-reported numbers are thus inflated. 
            \item Moreover, we permuted vertex indices in all input triangulations to prevent models from exploiting a combinatorial shortcut. This resulted in inflated predictive performance in the previous version of this work and has since been rectified.
\item Previously, we assigned the $\mathbf{R}$ encoding to vertices, i.e., $0$-simplices, and then assigned the barycenter of those to each $k \geq 1$-simplex. To generate truly random features, we now assign random features to \emph{all} simplices.
            \item We swept across an extra hyperparameter for dropout of $0.0$ to see if regularization was too strong in GCN and RG-GCN.
            \item We ran additional CWN trainings with an updated parameter count, i.e., $500$K, to make its size more aligned with the other models.
            \item We removed the specific performance of Graphormer on $3$D data to make space for more text and discussions.
            \item We added plots showing the out-of-distribution behavior of models for the $2$D-balanced and $3$D settings.
            \item We added additional comments that interpret the results.
        \end{compactitem} 
    \item Changes in the \emph{appendix}:
        \begin{compactitem}
            \item We updated the oversampling procedure for deduplication.
            \item We removed redundant tables and added heatmaps for all configurations instead.
        \end{compactitem}
\end{compactitem}

We very much appreciate the feedback by reviewers, which helped us substantially improve our work.

\section{Model selection}

The full table of selection criteria considering HOMP models is \cref{tbl:full_eval_models} of which \cref{tbl:eval_models} is an abbreviated version for the main text.

\begin{table}[tbp]
    \centering
  \caption{Contenders for evaluations. The symbols denote the degree of attainment of a \emph{criterion}. The $\mathcolor{GHigh}{\checkmark}$ is complete,  $\mathcolor{OISkyBlue}{\sim}$ is partial, and $\mathcolor{OIBlue}{\times}$ indicates lack. The * marks a variation present in \texttt{TopoBench} and $\dagger$ a variation we benchmark.}

\begin{tabular}{llcccc}
    \toprule
     Domain &  Model & \textbf{R}elevant & \textbf{A}vailable & \textbf{V}erifiable & Comp. \textbf{E}ff.  \\
    \midrule
    \multirow{3}{*}{Graph} & GCN $(\dagger)$  & $\mathcolor{GHigh}{\checkmark}$ & $\mathcolor{GHigh}{\checkmark}$ & $\mathcolor{GHigh}{\checkmark}$ & $\mathcolor{GHigh}{\checkmark}$ \\
    & RG-GCN $(\dagger)$ & $\mathcolor{GHigh}{\checkmark}$ & $\mathcolor{GHigh}{\checkmark}$ & $\mathcolor{GHigh}{\checkmark}$ & $\mathcolor{GHigh}{\checkmark}$\\
    & Graphormer $(\dagger)$ & $\mathcolor{GHigh}{\checkmark}$ & $\mathcolor{GHigh}{\checkmark}$ & $\mathcolor{GHigh}{\checkmark}$ & $\mathcolor{GHigh}{\checkmark}$ \\
    \midrule
    \multirow{8}{*}{Simplicial complex} & SCNN & $\mathcolor{GHigh}{\checkmark}$ & $\mathcolor{OISkyBlue}{\sim}$ & $\mathcolor{GHigh}{\checkmark}$ & $\mathcolor{GHigh}{\checkmark}$ \\
    & SCCNN & $\mathcolor{GHigh}{\checkmark}$ & $\mathcolor{OIBlue}{\times}$ & $\mathcolor{OIBlue}{\times}$ & $\mathcolor{GHigh}{\checkmark}$\\
    & SCCNN (*) $(\dagger)$ & $\mathcolor{OISkyBlue}{\sim}$ & $\mathcolor{GHigh}{\checkmark}$ & $\mathcolor{OISkyBlue}{\sim}$ & $\mathcolor{GHigh}{\checkmark}$\\
    & SAN & $\mathcolor{GHigh}{\checkmark}$ & $\mathcolor{OISkyBlue}{\sim}$ & $\mathcolor{OISkyBlue}{\sim}$ &$\mathcolor{OIBlue}{\times}$\\
    & SAN (*) & $\mathcolor{OISkyBlue}{\sim}$ & $\mathcolor{GHigh}{\checkmark}$ & $\mathcolor{OIBlue}{\times}$ &$\mathcolor{GHigh}{\checkmark}$\\
    & CXN & $\mathcolor{OIBlue}{\times}$ & $\mathcolor{OIBlue}{\times}$ & $\mathcolor{OIBlue}{\times}$ &$\mathcolor{GHigh}{\checkmark}$\\
    & CIN & $\mathcolor{GHigh}{\checkmark}$ & $\mathcolor{OISkyBlue}{\sim}$ & $\mathcolor{GHigh}{\checkmark}$ & $\mathcolor{GHigh}{\checkmark}$\\
    & CWN (*)$(\dagger)$  & $\mathcolor{OISkyBlue}{\sim}$ & $\mathcolor{GHigh}{\checkmark}$ & $\mathcolor{OIBlue}{\times}$ & $\mathcolor{GHigh}{\checkmark}$\\
    & GCCN & $\mathcolor{GHigh}{\checkmark}$ & $\mathcolor{GHigh}{\checkmark}$ & $\mathcolor{GHigh}{\checkmark}$ & $\mathcolor{OIBlue}{\times}$\\
    & SMCN & $\mathcolor{OISkyBlue}{\sim}$ & $\mathcolor{GHigh}{\checkmark}$ & $\mathcolor{GHigh}{\checkmark}$ & $\mathcolor{OIBlue}{\times}$ \\
    & CT & $\mathcolor{GHigh}{\checkmark}$ & $\mathcolor{OIBlue}{\times}$ & $\mathcolor{OIBlue}{\times}$ & $\mathcolor{OIBlue}{\times}$ \\
    \bottomrule
  \end{tabular}

  \label{tbl:full_eval_models}
\end{table}

\section{Computational complexity}

\cref{tbl:comp_graph} and \cref{tbl:comp_sc} show the computational complexities of the graph models and HOMP models, respectively. 

\begin{table}[tbp]
    \centering
    \caption{Complexities of graph models. We denote the number of $k$-simplices by $V_k$.
    }
      \begin{tabular}{@{}lccc@{}}
        \toprule
         & $\mathcal{S}$ & $\mathcal{D}$ & $\mathcal{H}$ \\ 
        \midrule
        GCN & $\mathcal{O}(V_0 + V_1)$ & $\mathcal{O}(V_k + V_{k-1})$ & $\mathcal{O}\left(\sum_{i=0}^k V_i + \sum_{i=2}^k (i+1) \cdot V_i\right)$ \\
        RG-GCN & $\mathcal{O}(V_1)$ & $\mathcal{O}(V_{k-1})$ & $\mathcal{O}\left(\sum_{i=2}^k (i+1) \cdot V_i\right)$  \\
        Graphormer & $\mathcal{O}(V_0^2)$ & $\mathcal{O}(V_k^2)$ & $\mathcal{O}\left(\left(\sum_{i=0}^k V_i\right)^2 \right)$ \\
        \bottomrule
      \end{tabular}
      \label{tbl:comp_graph}
\end{table}

\begin{table}[tbp]
    \centering
    \caption{
        Complexities of HOMP models. The * denotes a \emph{conservative} estimate. In the case of GCCN, this estimate is obtained by restricting to \mbox{2-neighborhoods}. Entries with ``---'' indicate a missing theoretical complexity analysis, in which case we validated empirically that the models were not runnable. For a triangulation $\mathcal{T}$, we write $V_k$ for the number of $k$-simplices and $B_k$ for the number of \mbox{$(k-1)$-simplices} in the respective boundaries.
    }
      \begin{tabular}{@{}lccc@{}}
        \toprule
         & $\mathcal{T}$ \\ 
        \midrule
        SCCNN & $\mathcal{O}\left( (k+1) V_k + \sum_{i=0}^{k-1} (i+1) (V_i + V_{i+1}) \right)$  \\
        SAN & --- \\
        CIN & $\mathcal{O}\left( \sum_{i=1}^k B_i\cdot V_i + 2\cdot \binom{B_i}{2} V_i \right)$  \\
        CWN & $\mathcal{O}\left( (k+1) V_k + \sum_{i=0}^{k-1} (i+1) (V_i + V_{i+1}) \right)$  \\
        GCCN (*) & $\mathcal{O}\left(V_0 + V_1 + V_1  + V_2  + \sum_{i=1}^{k} V_{i-1} \cdot V_i \right)$& \\
        SMCN & $\mathcal{O}\left(\max \left(\{B_i \mid i = 0, \dots k-1\} \cdot V_0 \cdot V_1 \right) \right)$\\
        CT & --- \\
        \bottomrule
      \end{tabular}
      \label{tbl:comp_sc}
\end{table}

\section{Hardware specifications}
\label{app:hardware_config}

For model training, we use $4$ \textsc{AMD EPYC 9654 96-Core Processor} and $8$ \textsc{NVIDIA RTX PRO 6000 Blackwell Server Edition} with \texttt{CUDA} version 13.0, and \texttt{PyTorch} version 2.9.0.
For dataset precomputations, we use a compute node with $\times 2$ CPU \textsc{AMD EPYC 7763 64-Core Processor} and  1 TiB \textsc{DRAM}. 

\section{Hyperparameter configurations}\label{section:app_hyperparams}

We report our general hyperparameters in \cref{tbl:hyperparams_general} and model-specific hyperparameters in \Cref{tab:Specific hyperparameters}.

\begin{table}[tbp]
    \centering
    \caption{General hyperparameters}
\label{tbl:hyperparams_general}
\begin{tabular}{l c }
    \toprule
    Hyperparameter & Values  \\
    \midrule
    Learning rate &  $\{0.01, 0.001\}$ \\
    Batch size &  $\{128, 256\}$ \\
    Max epochs & $\{300\}$ \\
    Early stopping & $\{100\}$ \\
    Optimizer & Adam \\
    Seeds & $\{41, 42, 43\}$ \\
    Train/Val/Test-Split & 60/20/20 \\
    \bottomrule
\end{tabular} \end{table}

\begin{table}[t]
\centering
\caption{
Model hyperparameters. $^\dagger$ denotes the per-cell dimension, while SPD denotes the \emph{shortest-path distance}.
}
\label{tab:Specific hyperparameters}
\begin{tabular}{l ccc cc}
\toprule
 & \multicolumn{3}{c}{Graph} & \multicolumn{2}{c}{Higher-order} \\
\cmidrule(lr){2-4}\cmidrule(lr){5-6}
Hyperparameter & GCN & RG-GCN & Graphormer & SCCNN & CWN \\
\midrule
\multicolumn{6}{l}{\emph{Shared}} \\
\midrule
Hidden dimension     & $\{96,172\}$ & $\{80,96\}$ & $\{48,64\}$ & $64^\dagger$ & $\{ 64, 128 \}^\dagger$ \\
Number of layers     & $\{8,16\}$ & $\{6,14\}$ & $\{4,8\}$ & $\{3,\dots,6\}$ & $\{3,\dots,6\}$ \\
Activation           & ReLU & ReLU & GELU & ReLU & ELU \\
Normalization        & Batch & Batch & Layer & --- & --- \\
Residual             & $\checkmark$ & $\checkmark$ & $\checkmark$ & --- & --- \\
Dropout              & $\{0,0.5\}$ & $\{0,0.5\}$ & $0.1$ & --- & --- \\
Readout              & Mean & Mean & Mean & Sum & Sum \\
\midrule
\multicolumn{6}{l}{\emph{Model-specific}} \\
\midrule
Residual gate activation  & & Sigmoid & & & \\
Number of heads           & & & 8 & & \\
Spatial encoding          & & & SPD & & \\
Max degree / distance     & & & 32 / 16 & & \\
Conv.\ / complex order    & & & & 1 / 2 & \\
Aggregation normalization & & & & \texttimes & \\
Aggregation / update      & & & & & Sum / ELU \\
\bottomrule
\end{tabular}
\end{table}

\paragraph{Encoding hyperparameters.}
For the encodings, we chose constant hyperparameters across all models to assure comparability, in particular, the random walk was always performed with $8$ steps, the random features had dimension $8$. The moment curve embedding and node degree encoding do not contain hyperparameters.

\paragraph{Data balancing hyperparameters.}
The data balancing was run with a target count of $2,500$ and $5,000$ samples per class for the $2$D and $3$D datasets, respectively. The maximal subgroup size, until which direct isomorphism checks were performed, was $5$. We first oversampled the required samples by a factor of $2$ and afterwards applied the deduplication procedure.

\begin{figure}[tbp]
\centering
\raisebox{0.125\textwidth}{\rotatebox[origin=c]{90}{\textbf{GCN}}}\hfill
    \subcaptionbox{$2$D-unbalanced}{
        \includegraphics[width=0.3\textwidth]{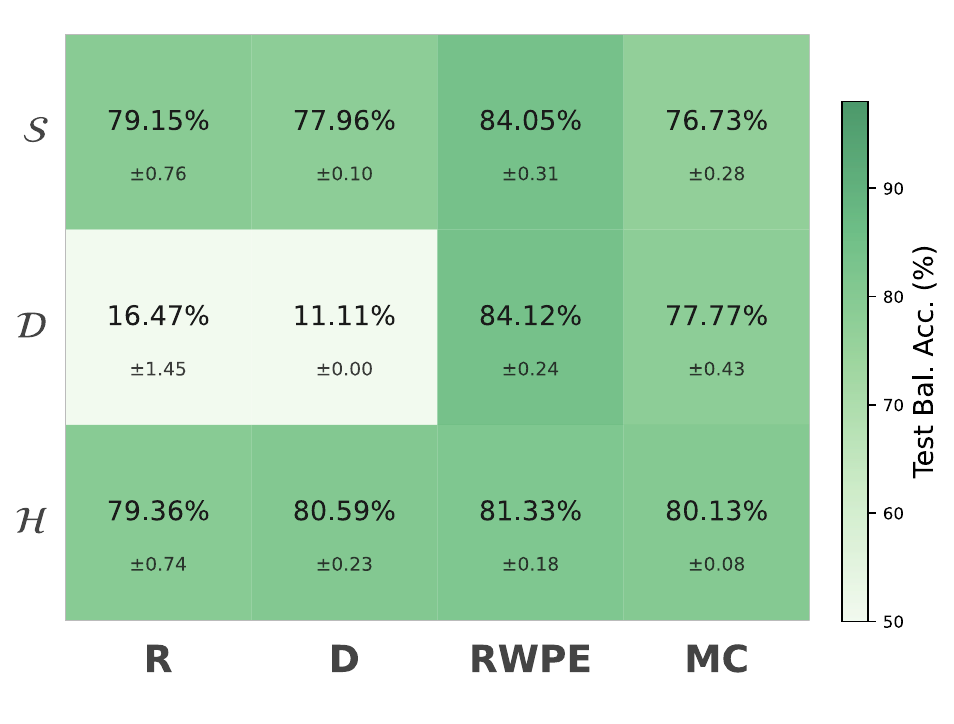}        
    }
    \subcaptionbox{$2$D-balanced}{
        \includegraphics[width=0.3\textwidth]{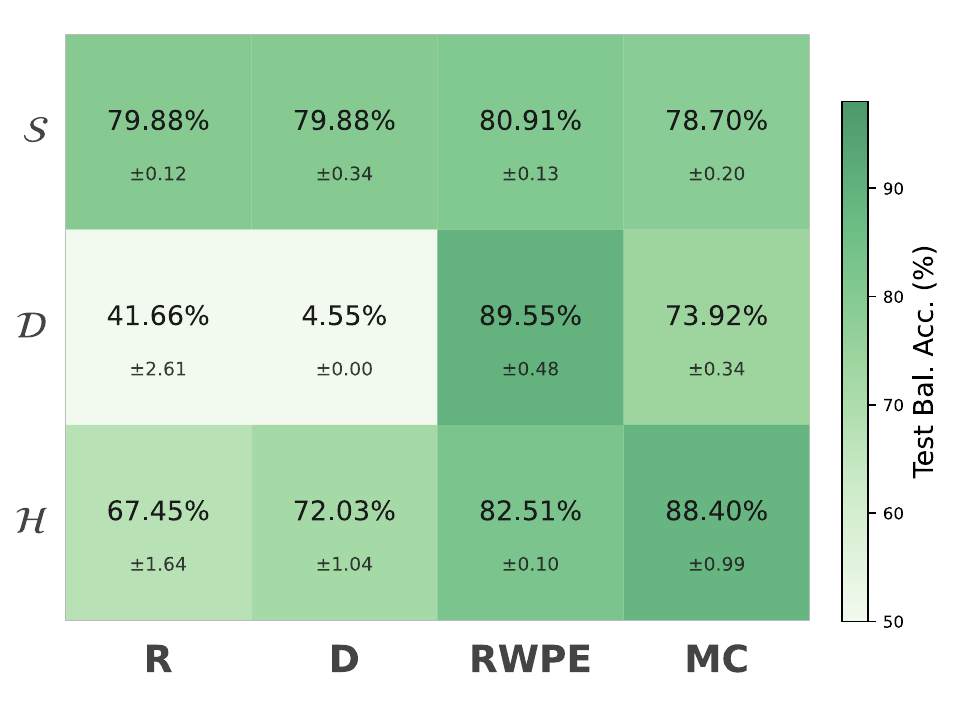}
    }
    \subcaptionbox{$3$D-balanced}{
        \includegraphics[width=0.3\textwidth]{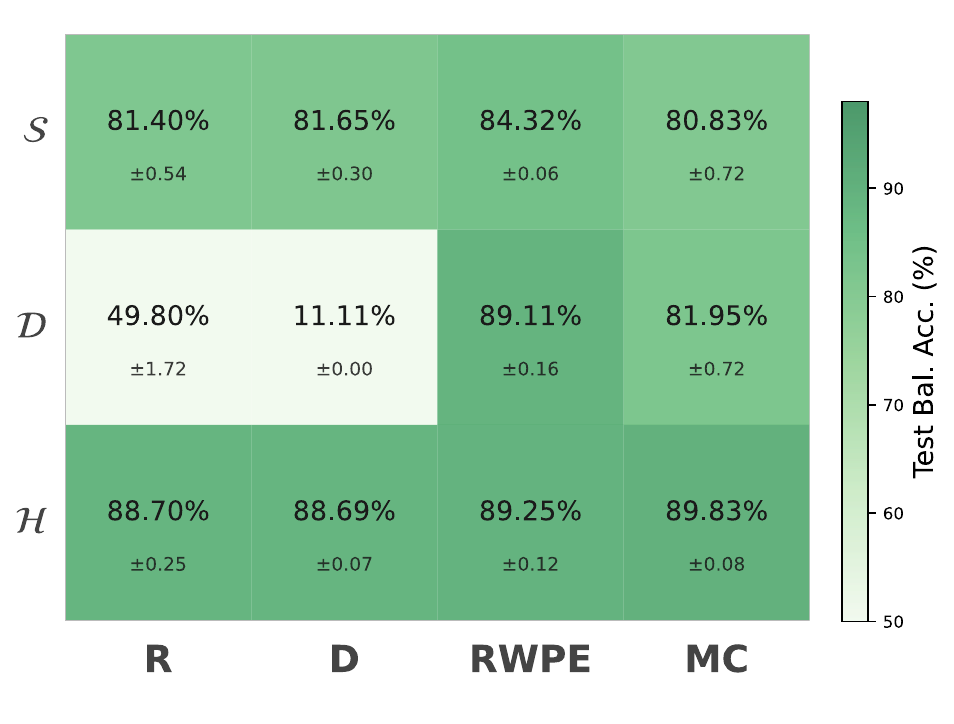}
    }    

\raisebox{0.125\textwidth}{\rotatebox[origin=c]{90}{\textbf{RG-GCN}}}\hfill
    \subcaptionbox{$2$D-unbalanced}{
        \includegraphics[width=0.3\textwidth]{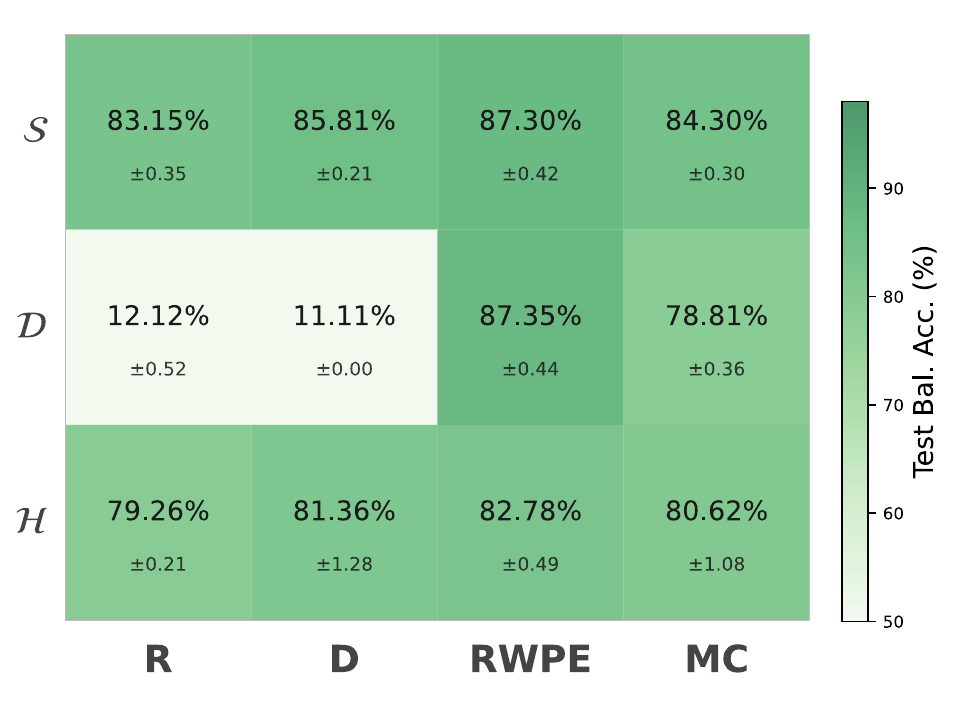}        
    }
    \subcaptionbox{$2$D-balanced}{
        \includegraphics[width=0.3\textwidth]{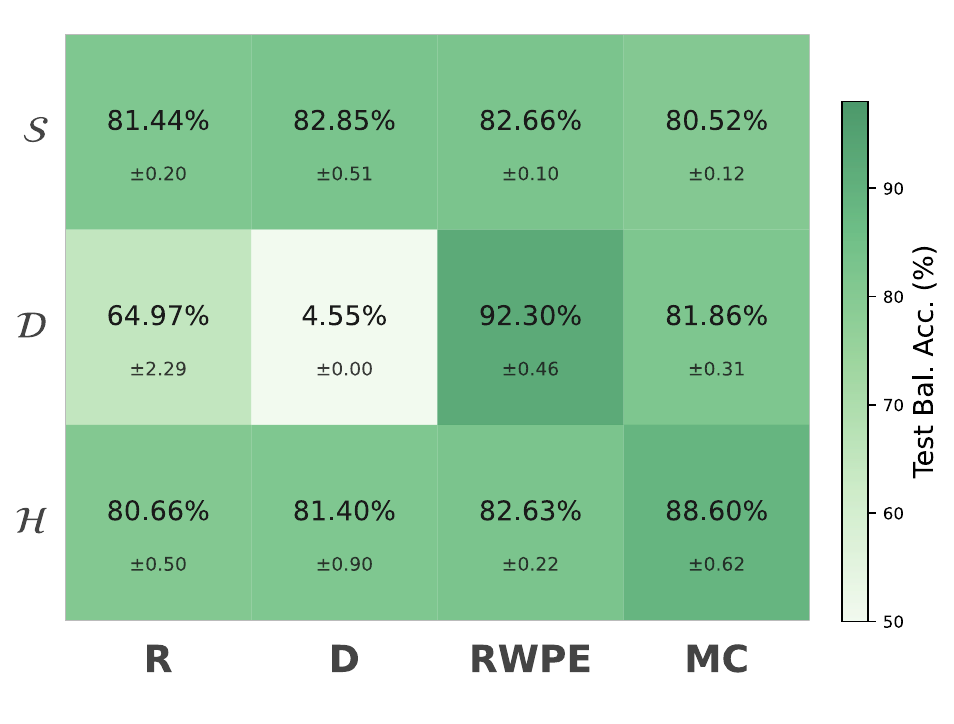}
    }
    \subcaptionbox{$3$D-balanced}{
        \includegraphics[width=0.3\textwidth]{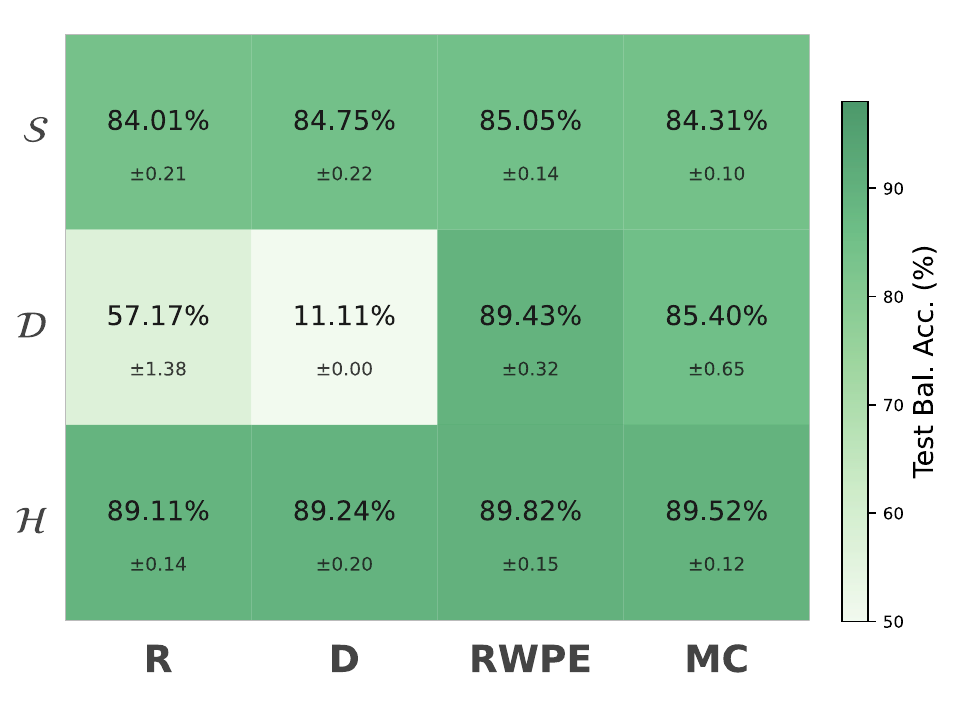}
    }     
    
\raisebox{0.125\textwidth}{\rotatebox[origin=c]{90}{\textbf{Graphormer}}}\hfill
    \subcaptionbox{$2$D-unbalanced}{
        \includegraphics[width=0.3\textwidth]{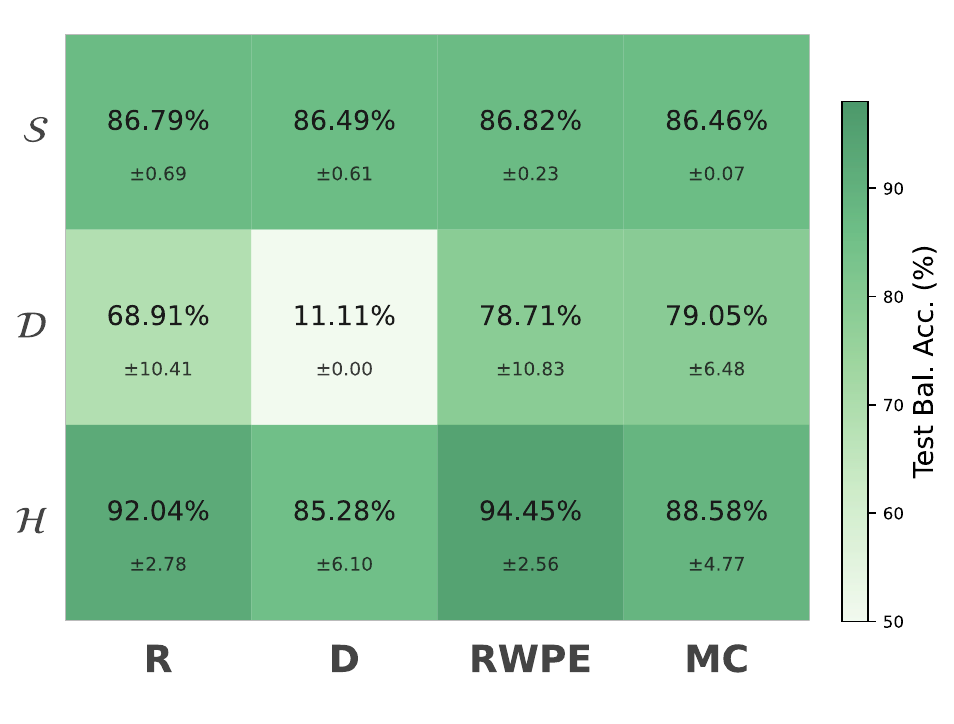}        
    }
    \subcaptionbox{$2$D-balanced}{
        \includegraphics[width=0.3\textwidth]{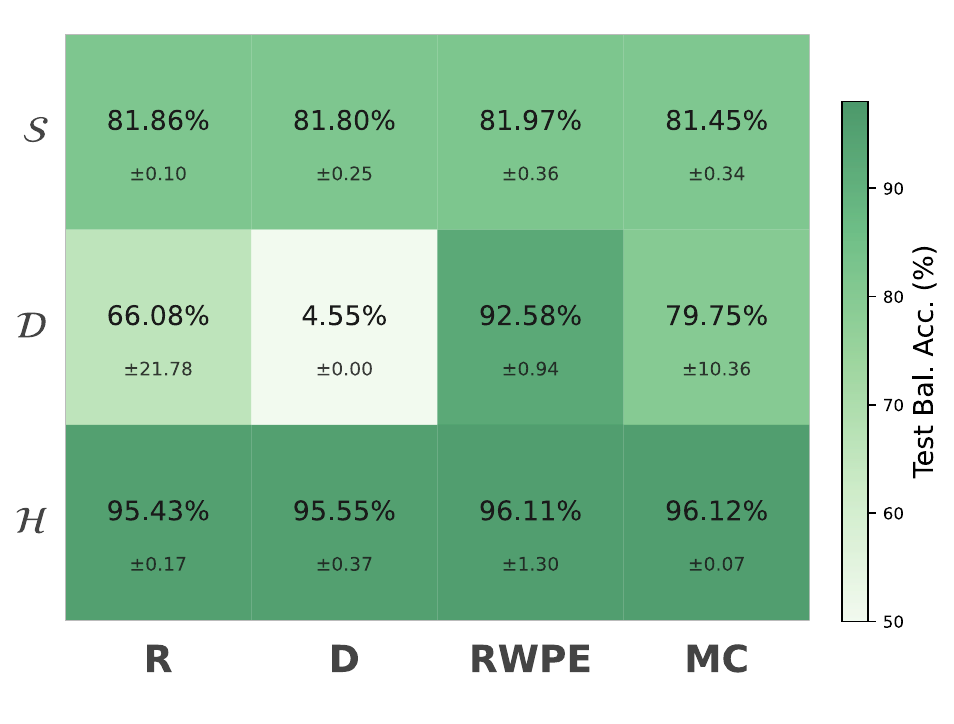}
    }
    \subcaptionbox{$3$D-balanced}{
        \includegraphics[width=0.3\textwidth]{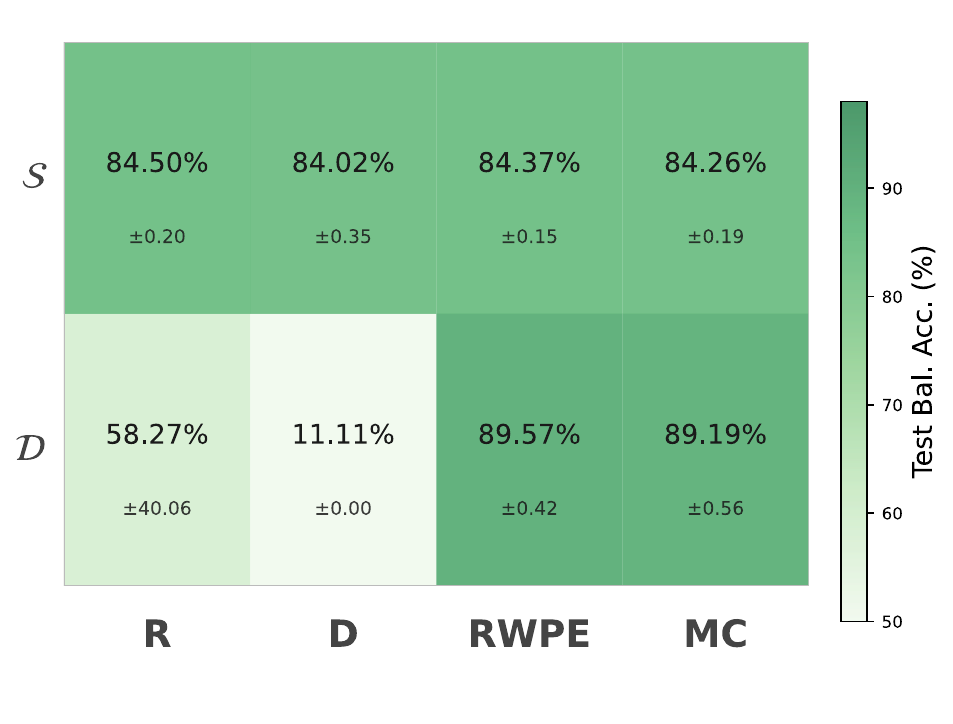}
    }
    
\raisebox{0.125\textwidth}{\rotatebox[origin=c]{90}{\textbf{SCCNN}}}\hfill
    \subcaptionbox{$2$D-unbalanced}{
        \includegraphics[width=0.3\textwidth]{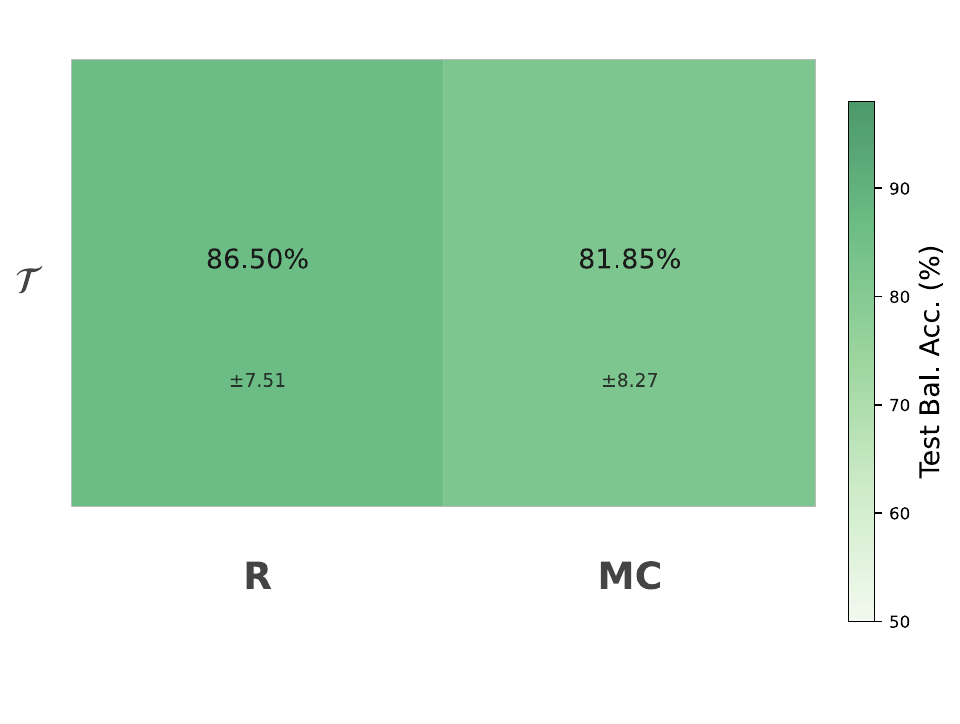}        
    }
    \subcaptionbox{$2$D-balanced}{
        \includegraphics[width=0.3\textwidth]{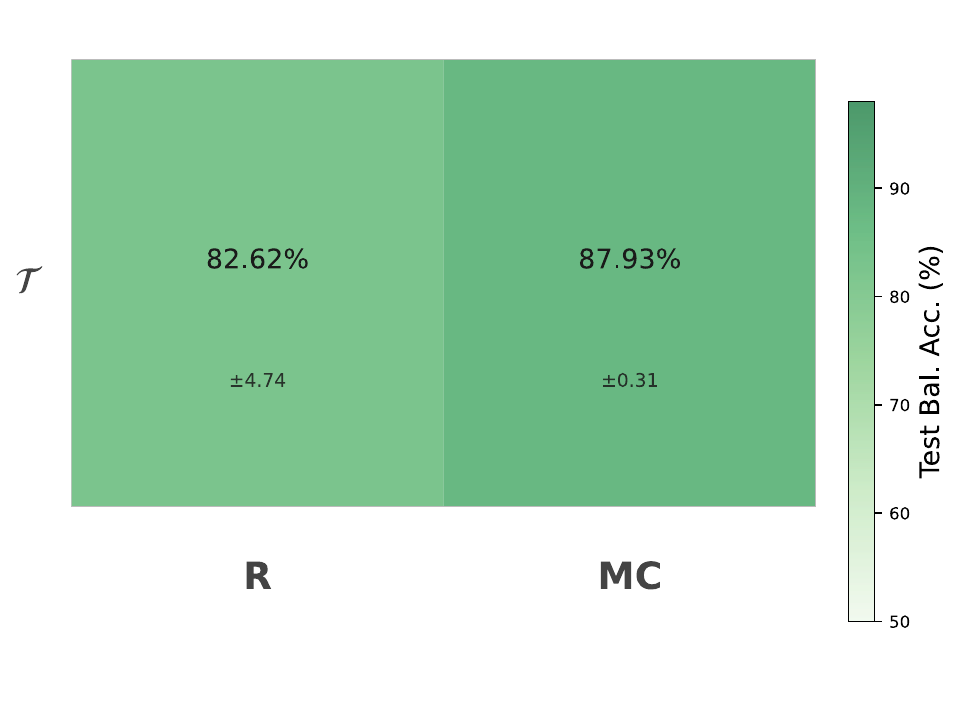}
    }
    \subcaptionbox{$3$D-balanced}{
        \includegraphics[width=0.3\textwidth]{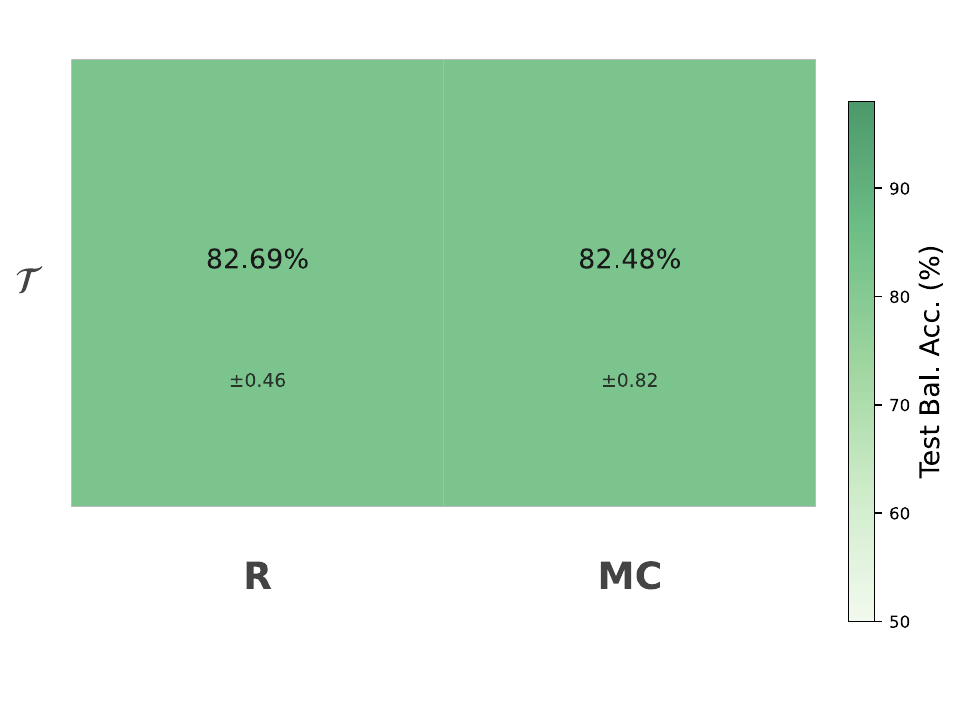}
    }

\raisebox{0.125\textwidth}{\rotatebox[origin=c]{90}{\textbf{CWN}}}\hfill
    \subcaptionbox{$2$D-unbalanced}{
        \includegraphics[width=0.3\textwidth]{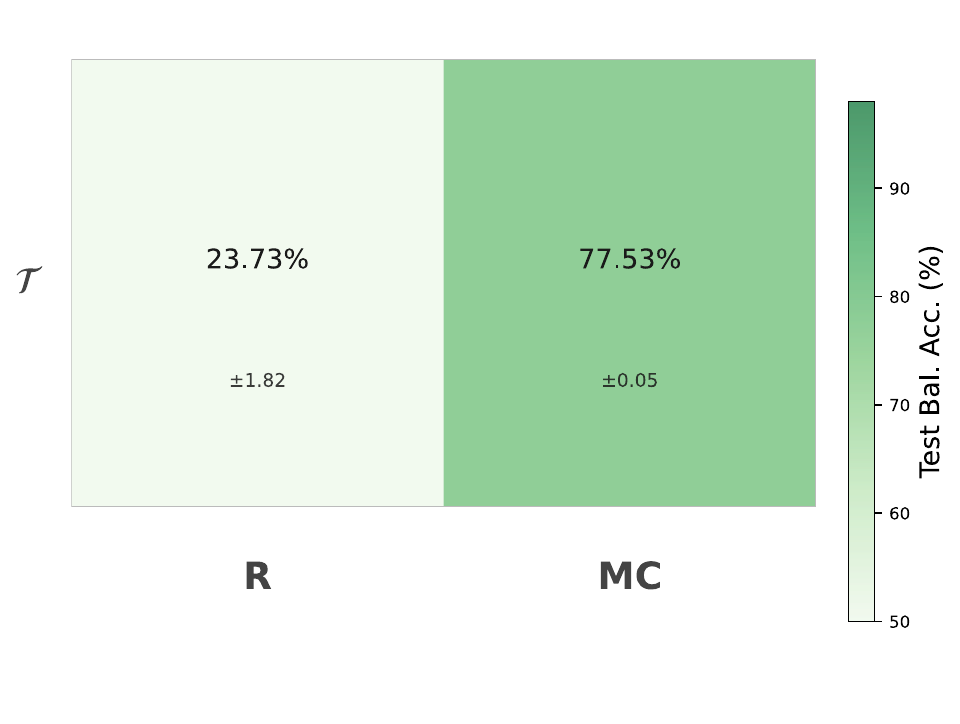}        
    }
    \subcaptionbox{$2$D-balanced}{
        \includegraphics[width=0.3\textwidth]{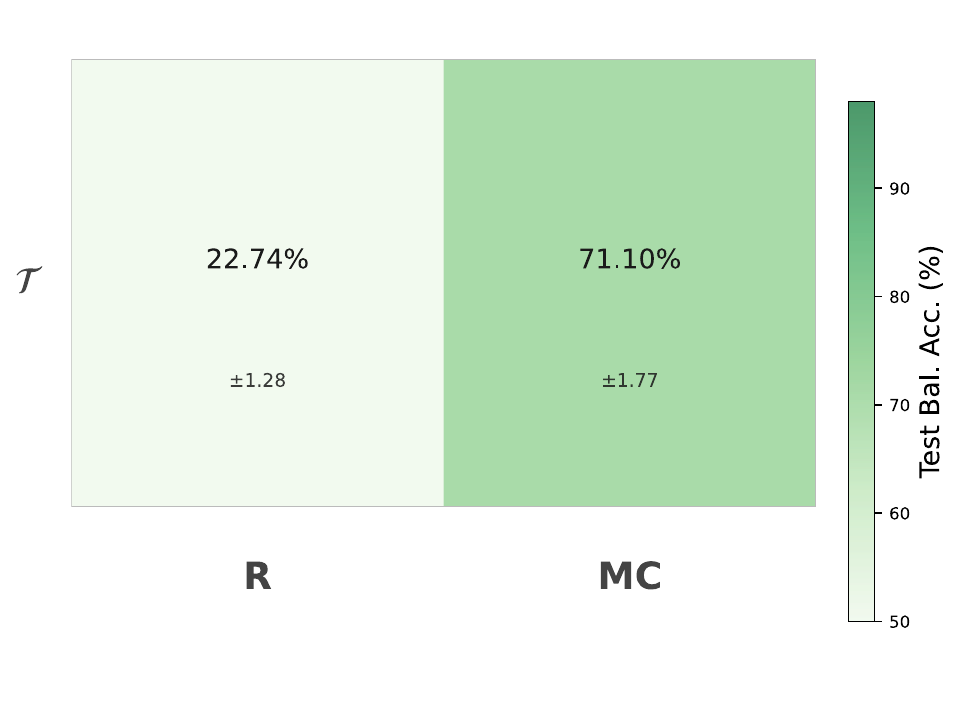}
    }
    \subcaptionbox{$3$D-balanced}{
        \includegraphics[width=0.3\textwidth]{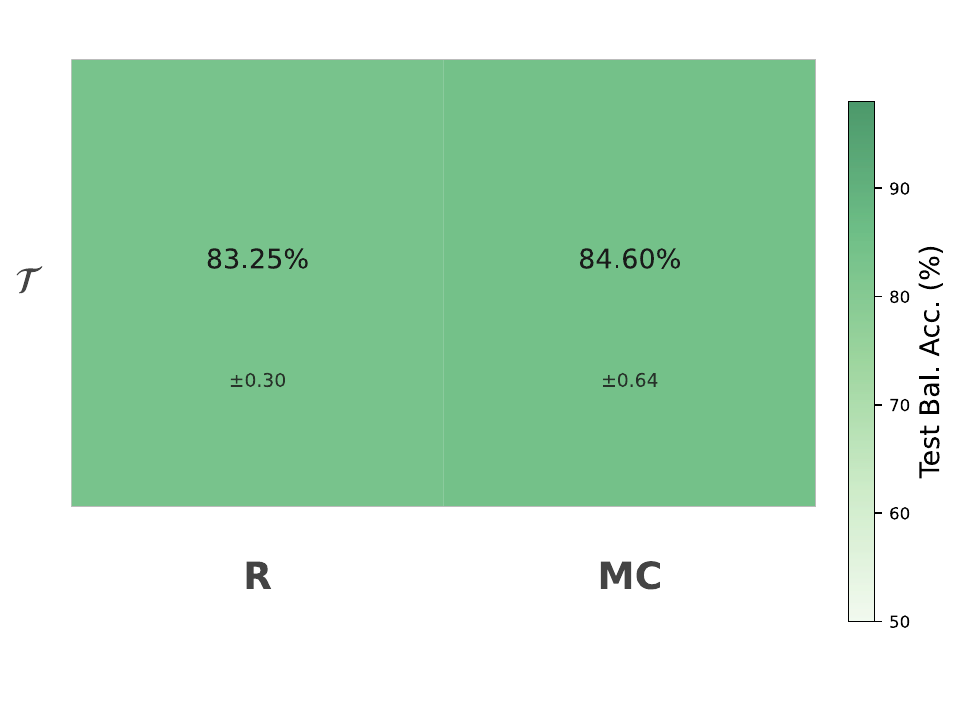}
    }
    \caption{Performance heatmaps for all models. The $x$-axis represents the encoding while the $y$-axis shows the representation.
    }
    \label{fig:Heatmaps}
\end{figure}

\section{Extended results}

\begin{figure}[tbp]
    \centering
    \includegraphics[width=0.95\linewidth,]{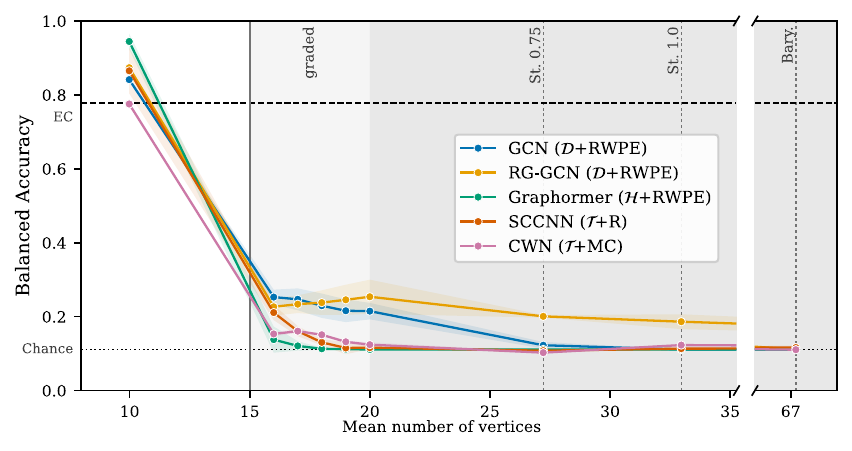}
    \caption{Balanced accuracy on $2$D-unbalanced under increasingly fine subdivisions. The leftmost point denotes the in-distribution test set while the shaded region covers the out-of-distribution subdivision datasets. Balanced accuracy drops below the Euler characteristic heuristic~(EC, dashed) after a \emph{single} graded subdivision and rapidly decays to chance level~(dotted) as the number of subdivision steps increases.
    }
    \label{fig:ood_2d_ubal}
\end{figure}

\Cref{fig:Heatmaps} depicts  in-distribution results for all models, broken down by encoding and representation, respectively.
As described in \Cref{sec:topological-generalization}, we also evaluate all models on the stellar subdivision of $75 \%$ of the maximal simplices, on the stellar subdivision on all maximal simplices, and on the full barycentric subdivision. \Cref{fig:ood_2d_ubal} depicts out-of-distribution plots for the $2$D-unbalanced setting.

\section{Theory}

\begin{theorem}
    \label{theorem:class_2d}
    A closed connected $2$-manifold $\mathcal{M}$ is homeomorphic to one of the following:
    \begin{compactitem}
        \item The sphere.
        \item The connected sum of $k \geq 1$ tori, written as $\#^k\ T_2$.
        \item The connected sum of $k \geq 1$ real projective planes, written as $\#^k\ \mathbb{RP}_2$.
    \end{compactitem}
\end{theorem}
This theorem was proved multiple times, one of the first rigorous proofs was shown by Brahana in 1921 \cite{Brahana1921SystemsOC}. It allows classification of closed compact \textit{surfaces} by their invariants through this corollary.

\begin{corollary}
    Two closed compact surfaces $\mathcal{M}_1$ and $\mathcal{M}_2$ with $\dim \mathcal{M}_1=\dim \mathcal{M}_2 = 2$ are homeomorphic if and only if
\begin{inparaenum}[(i)]
        \item $\chi(\mathcal{M}_1) = \chi(\mathcal{M}_2)$, and
        \item they are either both orientable or both non-orientable,
    \end{inparaenum}
    where $\chi$ is the Euler characteristic.
    \label{cor:inv_class}
\end{corollary}  
\clearpage
\section*{NeurIPS Paper Checklist}

\begin{enumerate}

\item {\bf Claims}
    \item[] Question: Do the main claims made in the abstract and introduction accurately reflect the paper's contributions and scope?
    \item[] Answer: \answerYes{} \item[] Justification: The claims in the abstract match the content of the paper.

\item {\bf Limitations}
    \item[] Question: Does the paper discuss the limitations of the work performed by the authors?
    \item[] Answer: \answerYes{} \item[] Justification: We discuss the robustness of our evaluation method, the efficiency of our assumptions and finally the limitations throughout the work.

\item {\bf Theory assumptions and proofs}
    \item[] Question: For each theoretical result, does the paper provide the full set of assumptions and a complete (and correct) proof?
    \item[] Answer: \answerYes{} 
    \item[] Justification: The theoretical part is mostly relegated to the appendix and contains all the required assumptions and proofs.

    \item {\bf Experimental result reproducibility}
    \item[] Question: Does the paper fully disclose all the information needed to reproduce the main experimental results of the paper to the extent that it affects the main claims and/or conclusions of the paper (regardless of whether the code and data are provided or not)?
    \item[] Answer: \answerYes{}\item[] Justification: The results are all reproducible via the specified methods of generating the new datasets along with the code provided in the submission. Since the reference dataset is based on \texttt{MANTRA}, all of the processes thus described should be able to be done even without our specific code for them. Additionally, we attach to the submission our benchmarking suite that groups models and datasets for evaluation.

\item {\bf Open access to data and code}
    \item[] Question: Does the paper provide open access to the data and code, with sufficient instructions to faithfully reproduce the main experimental results, as described in supplemental material?
    \item[] Answer: \answerYes{}\item[] Justification: We provide a link to two repositories containing all the code and data generation information to faithfully reproduce the results.

\item {\bf Experimental setting/details}
    \item[] Question: Does the paper specify all the training and test details (e.g., data splits, hyperparameters, how they were chosen, type of optimizer) necessary to understand the results?
    \item[] Answer: \answerYes{}\item[] Justification: \cref{section:app_hyperparams} contains all hyperparameters and choices while Section 4 contains all other test details.

\item {\bf Experiment statistical significance}
    \item[] Question: Does the paper report error bars suitably and correctly defined or other appropriate information about the statistical significance of the experiments?
    \item[] Answer: \answerYes{}{} \item[] Justification: We report error bars or clearly state the error in each reported result.

\item {\bf Experiments compute resources}
    \item[] Question: For each experiment, does the paper provide sufficient information on the computer resources (type of compute workers, memory, time of execution) needed to reproduce the experiments?
    \item[] Answer: \answerYes{} \item[] Justification: We provide a section with hardware specifications of what was used during the experimental setup in \cref{app:hardware_config}.
    
\item {\bf Code of ethics}
    \item[] Question: Does the research conducted in the paper conform, in every respect, with the NeurIPS Code of Ethics \url{https://neurips.cc/public/EthicsGuidelines}?
    \item[] Answer: \answerYes{} \item[] Justification: We adhere to the code of ethics.

\item {\bf Broader impacts}
    \item[] Question: Does the paper discuss both potential positive societal impacts and negative societal impacts of the work performed?
    \item[] Answer: \answerNA{}{} \item[] Justification: No societal impact beyond the general possible impact of artificial intelligence systems.
    
\item {\bf Safeguards}
    \item[] Question: Does the paper describe safeguards that have been put in place for responsible release of data or models that have a high risk for misuse (e.g., pre-trained language models, image generators, or scraped datasets)?
    \item[] Answer: \answerNA{} \item[] Justification: No risk of misuse beyond general artificial intelligence misuse.

\item {\bf Licenses for existing assets}
    \item[] Question: Are the creators or original owners of assets (e.g., code, data, models), used in the paper, properly credited and are the license and terms of use explicitly mentioned and properly respected?
    \item[] Answer: \answerYes{}\item[] Justification: We cite the original MANTRA dataset \cite{ballester2025mantra} and mention its BSD-3-Clause license. We also cite all models used \cite{kipf2017semisupervised, bresson2017residual,ying2021graphormer, yang2025hodgeaware,bodnar2021cwnetworks} and their implementations \cite{telyatnikov2025topobench}. 

\item {\bf New assets}
    \item[] Question: Are new assets introduced in the paper well documented and is the documentation provided alongside the assets?
    \item[] Answer: \answerYes{} \item[] Justification: We correctly document the new assets introduced.

\item {\bf Crowdsourcing and research with human subjects}
    \item[] Question: For crowdsourcing experiments and research with human subjects, does the paper include the full text of instructions given to participants and screenshots, if applicable, as well as details about compensation (if any)? 
    \item[] Answer: \answerNA{} \item[] Justification: No crowdsourcing or human subjects.

\item {\bf Institutional review board (IRB) approvals or equivalent for research with human subjects}
    \item[] Question: Does the paper describe potential risks incurred by study participants, whether such risks were disclosed to the subjects, and whether Institutional Review Board (IRB) approvals (or an equivalent approval/review based on the requirements of your country or institution) were obtained?
    \item[] Answer: \answerNA{} \item[] Justification: No crowdsourcing or human participants.

\item {\bf Declaration of LLM usage}
    \item[] Question: Does the paper describe the usage of LLMs if it is an important, original, or non-standard component of the core methods in this research? Note that if the LLM is used only for writing, editing, or formatting purposes and does \emph{not} impact the core methodology, scientific rigor, or originality of the research, declaration is not required.
\item[] Answer: \answerNA{} \item[] Justification: Not used.

\end{enumerate} 
\end{document}